\documentclass[10pt,twocolumn,letterpaper]{article}

\usepackage{iccv}
\usepackage{times}
\usepackage{epsfig}
\usepackage{graphicx}
\usepackage{amsmath}
\usepackage{amssymb}
\usepackage[dvipsnames]{xcolor}
\usepackage[sort,numbers]{natbib}
\usepackage{booktabs}
\usepackage{multirow}
\usepackage{multicol}
\usepackage{sidecap}
\usepackage{stfloats}

\newcommand{\bz}{\boldsymbol{z}}
\newcommand{\bzt}{\boldsymbol{\tilde{z}}}
\newcommand{\bw}{\boldsymbol{w}}
\newcommand{\bwt}{\boldsymbol{\tilde{w}}}
\newcommand{\bmu}{\boldsymbol{\mu}}
\newcommand{\bx}{\boldsymbol{x}}
\newcommand{\norm}[1]{\left\lVert#1\right\rVert}

\definecolor{acg}{RGB}{255, 50, 50}

\definecolor{standard}{RGB}{247, 37, 133}
\definecolor{hyperbolic}{RGB}{58, 12, 163}
\definecolor{spherical}{RGB}{61, 161, 192}
\definecolor{arcface}{RGB}{61, 161, 192}
\definecolor{vmf}{RGB}{61, 161, 192}

\newcommand{\standard}{\textcolor{standard}{\textsc{standard}}}

\newcommand{\hyperbolic}{\textcolor{hyperbolic}{\textsc{hyperbolic}}}

\newcommand{\spherical}{\textcolor{spherical}{\textsc{cosine}}}

\newcommand{\arcface}{\textcolor{arcface}{\textsc{arcface}}}

\newcommand{\vmf}{\textcolor{vmf}{\textsc{vMF}}}

\usepackage[pagebackref=true,breaklinks=true,letterpaper=true,colorlinks,bookmarks=false]{hyperref}

\iccvfinalcopy 

\def\iccvPaperID{9099} 

\ificcvfinal\fi

\begin{document}

\title{von Mises--Fisher Loss:\\An Exploration of Embedding Geometries for Supervised Learning}

\author{Tyler R.~Scott\thanks{Work performed during an internship at Google Research.} \\
University of Colorado, Boulder \\
{\tt\small tysc7237@colorado.edu}
\and
Andrew C.~Gallagher  \\
Google Research \\
{\tt\small agallagher@google.com} \\
\and
Michael C.~Mozer \\
Google Research \\
{\tt\small mcmozer@google.com}
}

\maketitle
\ificcvfinal\thispagestyle{empty}\fi

\begin{abstract}
    
Recent work has argued that classification losses utilizing softmax cross-entropy are superior not only for fixed-set classification tasks, but also by outperforming losses developed specifically for open-set tasks including few-shot learning and retrieval. 
Softmax classifiers have been studied using different embedding geometries---Euclidean, hyperbolic, and spherical---and claims have been made about the superiority of one or another, but they have not been systematically compared with careful controls. 
We conduct an empirical investigation of embedding geometry on softmax losses for a variety of fixed-set classification and image retrieval tasks. 
An interesting property observed for the spherical losses lead us to propose a probabilistic classifier based on the von Mises--Fisher distribution, and we show that it is competitive with state-of-the-art methods while producing improved out-of-the-box calibration. 
We provide guidance regarding the trade-offs between losses and how to choose among them.
    
\end{abstract}

\section{Introduction}
\label{sec:intro}

Almost on a weekly basis, novel loss functions are proposed that claim superiority over standard losses for supervised learning in vision.
At a coarse level, these loss functions can be divided into \emph{classification based} and \emph{similarity
based}. 
Classification-based losses 
\cite{Bridle1990,Chang2020,Collier2020,Brebisson2016,Deng2019,Ganea2018,Liu2017,Movshovitz-Attias2017,Qian2019,Ranjan2019,Ranjan2017,Sun2020,Teh2020,Tian2020,Wang2018,Wang2018a,Wang2017normface,Zhai2019,Mettes2019}
have generally been applied to \emph{fixed-set} classification tasks (i.e., tasks in which the 
the set of classes in training and testing is identical).
The prototypical classification-based loss uses a softmax function to map an embedding to a probability
distribution over classes, which is then evaluated with cross-entropy \cite{Bridle1990}.
Similarity-based losses
\cite{Allen2019,Chopra2005,Fort2017,Hadsell2006,Kaiser2017,Khrulkov2020,Koch2015,Oh2019,Ridgeway2018,Rippel2016,Schroff2015,Snell2017,Sohn2016,Song2016,Sung2018,Ustinova2016,Vinyals2016,Wang2017,Wang2019,Zhang2017,Weinberger2009,Wu2017,Yuan2019,Hu2014,Li2014,Liu2015,Goldberger2004,Nickel2017,Cakir2019,Song2017,Trian2017,Wang2019ranked,Law2019}
have been designed specifically for \emph{open-set} tasks, which include retrieval and few-shot learning. 
Open-set tasks refer to situations in which the classes at testing are disjoint from, or sometimes a superset of, those available at training.
The prototypical similarity-based method is the triplet loss which discovers embeddings such that an 
instance is closer to instances of the same class than to instances of different classes
\cite{Weinberger2009,Schroff2015}.

Recent efforts to systematically compare losses support a provocative hypothesis: on open-set tasks, classification-based losses outperform similarity-based losses by leveraging embeddings in the layer immediately preceding the logits
\cite{Musgrave2020,Boudiaf2020,Tian2020,Zhai2019,Sun2020}.
The apparent advantage of classifiers stems from the fact that similarity losses require sampling
informative pairs, triplets, quadruplets, or batches of instances in order to train effectively 
\cite{Boudiaf2020,Deng2019,Movshovitz-Attias2017,Wang2018,Wang2017normface,Wang2018a,Zhai2019}.
However, all classification losses are not equal, and we find systematic differences among them with regard to a fundamental choice:
the \emph{embedding geometry}, which determines the similarity structure of the embedding space.

Classification losses span three embedding geometries: \emph{Euclidean}, \emph{hyperbolic}, and \emph{spherical}. Although some comparisons have been made between
geometries, the comparisons have not been entirely systematic and have not covered the variety of supervised tasks. We find this fact somewhat surprising given the many
large-scale comparisons of loss functions.
Furthermore, the comparisons that have been made appear to be contradictory.
The face verification community has led the push for spherical losses, claiming superiority of spherical over Euclidean. However, this work is limited to open-set face-related tasks \cite{Deng2019,Liu2017,Ranjan2017,Ranjan2019,Wang2018,Wang2017normface,Wang2018a}.
The deep metric-learning community has recently refocused its attention to classification losses, but it is unclear from empirical comparisons whether the best-performing geometry is Euclidean or spherical \cite{Zhai2019,Qian2019,Boudiaf2020,Musgrave2020}.
Independently, Khrulkov \etal~\cite{Khrulkov2020} show that a hyperbolic prototypical network is a strong performer on common few-shot learning benchmarks, and additionally a hyperbolic softmax classifier outperforms the Euclidean variant on person re-identification.
Unfortunately, these results are in contention with Tian \etal~\cite{Tian2020}, where the authors claim a simple Euclidean softmax classifier learns embedding that are superior for few-shot learning.

One explanation for the discrepant claims are confounds that make it impossible to determine whether the causal factor for the superiority
of one loss over another is embedding geometry or some other ancillary aspect of the loss. Another explanation is that each bit of research
examines only a subset of losses or a subset of datasets.
Also, as pointed out in Musgrave \etal~\cite{Musgrave2020}, experimental setups (e.g., using the test set as a validation signal, insufficient hyperparameter tuning, varying forms of data augmentation) make it difficult to trust and reproduce published results.
The goal of our work is to take a step toward rigor by reconciling differences among classification losses on both fixed-set and image-retrieval benchmarks.

As discussed in more detail in Section \ref{sec:spherical}, our investigations led us to uncover an interesting property of spherical losses, 
which in turn suggested a probabilistic spherical classifier based on the von Mises--Fisher distribution.
While our loss is competitive with state-of-the-art alternatives and produces improved out-of-the-box calibration, we avoid unequivocal claims about its superiority.
We do, however, believe that it improves on previously proposed stochastic classifiers (e.g., \cite{Scott2019,Oh2019}), in, for example, its ability to scale to higher-dimensional embedding spaces.

\paragraph{Contributions.} In our work, 
(1) we characterize classification losses in terms of embedding geometry,
(2) we systematically compare classification losses in a well-controlled setting
on a range of fixed- and open-set tasks, examining both accuracy and calibration,
(3) we reach the surprising conclusion that spherical losses generally outperform
the standard softmax cross-entropy loss that is used almost exclusively in
practice,
(4) we propose a stochastic spherical loss based on von Mises--Fisher distributions,
scale it to larger tasks and representational spaces than previous stochastic
losses, and show that it can obtain state-of-the-art performance with significantly
lower calibration error,
and
(5) we discuss trade-offs between losses and factors to consider when choosing among them.

\section{Classification Losses}
\label{sec:losses}

We consider classification losses that compute the cross-entropy between a predicted class distribution and a one-hot target distribution (or equivalently,
as the negative log-likelihood under the model of the target class). 
The geometry determines the specific mapping from a deep embedding to a class posterior, and in a classification loss, this mapping is determined by a set of parameters learned via gradient descent.
We summarize the three embedding geometries that serve to differentiate classification losses.

\subsection{Euclidean}

Euclidean embeddings lie in an $n$-dimensional real-valued space (i.e., $\mathbb{R}^n$ or sometimes $\mathbb{R}^n_+$).
The commonly-used dot-product softmax \cite{Bridle1990}, which we refer to as
\standard, has the form:

\begin{equation}
    p(y \vert \bz) = \frac{\exp(\bw_y^{\text{T}} \bz)}{\sum_j \exp(\bw_j^{\text{T}} \bz)},
\end{equation}
where $\bz$ is an embedding and $\bw_j$ are weights for class $j$.
The dot product is a measure of similarity in Euclidean space, and is related to
the Euclidean distance by $||\bw_j-\bz||^2 = ||\bw_j||^2 + ||\bz||^2 - 2 \bw_j^\text{T} \bz$. (Classifiers using Euclidean distance have been explored, but gradient-based
training methods suffer from the curse of dimensionality because gradients go to
zero when all points are far from one another. Prototypical networks \cite{Snell2017}
do succeed using a Euclidean distance posterior, but the weights are determined
by averaging instance embeddings, not gradient descent.)

\subsection{Hyperbolic}

We follow Ganea \etal~\cite{Ganea2018} and Khrulkov \etal~\cite{Khrulkov2020} and consider the Poincar\'e ball model of hyperbolic geometry defined as $\mathbb{D}^n_c = \{\bz \in \mathbb{R}^n : c\norm{\bz}^2 < 1, c \ge 0\}$ where $c$ is a hyperparameter controlling the curvature of the ball.
Embeddings thus lie inside a hypersphere of radius $1 / \sqrt{c}$.
To perform multi-class classification, we employ the hyperbolic softmax generalization derived in \cite{Ganea2018}, hereafter \hyperbolic:

\begin{equation}
\begin{split}
    p&(y \vert \bz) \propto \\
    &\exp\left(\frac{\lambda^c_{\boldsymbol{p}_y} \norm{\boldsymbol{a}_y}}{\sqrt{c}} \sinh^{-1}\left(\frac{2 \sqrt{c} \langle -\boldsymbol{p}_y \oplus_c \bz, \boldsymbol{a}_y \rangle}{\left(1 - c\norm{-\boldsymbol{p}_y \oplus_c \bz}^2\right) \norm{\boldsymbol{a}_y}}\right)\right),
\end{split}
\end{equation}

\noindent where $\boldsymbol{p}_j \in \mathbb{D}_c^n$ and $\boldsymbol{a}_j \in T_{\boldsymbol{p}_j} \mathbb{D}_c^n \hspace{0.02in} \setminus \hspace{0.02in} \{\boldsymbol{0}\}$\footnote{$T_{\boldsymbol{x}} \mathbb{D}_c^n$ denotes the tangent space of $\mathbb{D}_c^n$ at $\boldsymbol{x}$.} are learnable parameters for class $j$, $\lambda^c_{\boldsymbol{p}_j}$ is the conformal factor of $\boldsymbol{p}_j$, $\langle . \rangle$ is the dot product, and $\oplus_c$ is the M\"obius addition operator. Further details can be found in \cite{Ganea2018,Khrulkov2020}.

\subsection{Spherical}
\label{sec:spherical}

Spherical embeddings lie on the surface of an $n$-dimensional unit-hypersphere (i.e., $\mathbb{S}^{n-1}$). 
The traditional loss, hereafter \spherical, uses cosine similarity \cite{Zhai2019}:

\begin{equation}
    p(y \vert \bz) = \frac{\exp(\beta \cos \theta_y)}{\sum_j \exp(\beta \cos \theta_j)},
\label{eq:spherical}
\end{equation}

\noindent where $\beta > 0$ is an inverse-temperature parameter, $\norm{\bz} = 1$, $\norm{\bw_j} = 1 \; \forall \; j$, and $\theta_j$ is the angle between $\bz$ and $\bw_j$.
Note that, in contrast to \standard, the $\ell_2$-norms are factored out of the weight vectors and embeddings, thus only the \emph{direction} determines class association.

Many variants of \spherical~have been proposed, particularly in the face verification community \cite{Deng2019,Liu2017,Ranjan2017,Ranjan2019,Wang2018,Wang2018a,Wang2017normface}, some of which are claimed to be superior.
For completeness, we also experiment with ArcFace \cite{Deng2019}, one of the top-performing variants, hereafter \arcface:

\vspace{-0.1in}

\begin{equation}
    p(y \vert \bz) = \frac{\exp(\beta \cos(\theta_y + m))}{\exp(\beta \cos(\theta_y + m)) + \sum_{j \ne y} \exp(\beta \cos \theta_j)},
\end{equation}

\noindent where $m \ge 0$ is an additive-angular-margin hyperparameter penalizing the true class.
(Note that we are coloring the loss name by geometry; \spherical\ and \arcface\ are both spherical losses.)

\begin{figure}[t]
\centering
\includegraphics[width=0.475\textwidth]{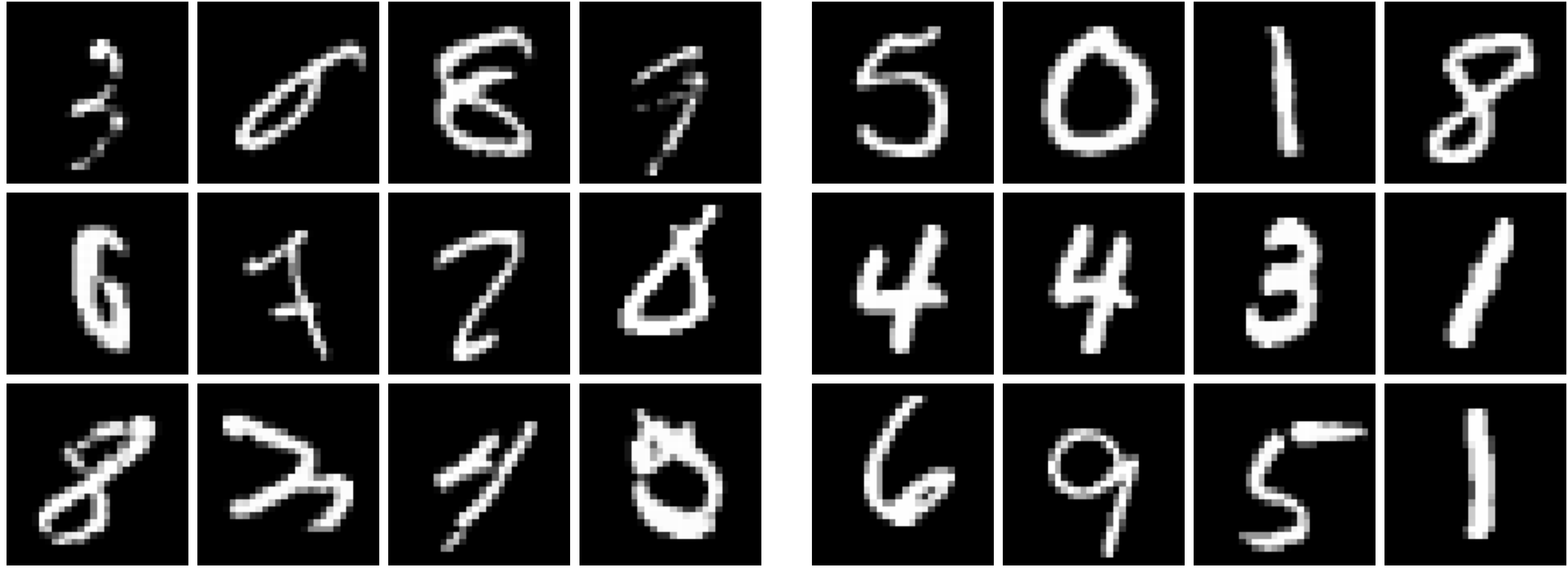}
\caption{MNIST test images corresponding to embeddings with the (left) smallest $\norm{\bz}$ and (right) largest $\norm{\bz}$; trained with \spherical. The left grid clearly contains ``noisier'' or unorthodox digits.}
\label{fig:spherical_mnist}
\end{figure}

Early in our investigations, we noticed an interesting property of spherical losses: 
$\norm{\bz}$ encodes information about uncertainty or ambiguity. For example, the left and right frames of Figure \ref{fig:spherical_mnist} 
show MNIST \cite{LeCun2010} test images that, when trained with \spherical, produce embeddings that have small and large
$\ell_2$-norms, respectively. 
This result is perfectly intuitive for \standard~since the norm affects the confidence
or peakedness of the class posterior distribution (verified in \cite{Ranjan2017,Parde2017}), but for \spherical, the norm has \emph{absolutely no effect} on the posterior.
Because the norm is factored out by the cosine similarity, there is no force on the model during training to reflect the ambiguity of an
instance in the norm.
Despite ignoring it, the \spherical\ model better discriminates correct versus incorrect predictions with the norm than
does the \standard\ model (see \spherical\ and \standard\ rows of Table~\ref{tab:auroc_correct}; note that the row for \vmf~corresponds to a loss we introduce in the next section).

Why does the \spherical\ embedding convey a confidence signal in the norm? One intuition is that when an instance is ambiguous, it
could be assigned many different labels in the training set, each pulling the instance's embedding in different directions. If these
directions roughly cancel, the embedding will be pulled to the origin.

Due to \spherical~having claimed advantages over \standard, and also discarding important information conveyed by $\norm{\bz}$,
we sought to develop a variant of \spherical\ that uses the $\ell_2$-norm to explicitly represent uncertainty in the embedding space, 
and thus to inform the classification decision. We refer to this variant as the \emph{von Mises--Fisher loss} or \vmf.

\begin{table}[t]
\small{
\begin{center}
\begin{tabular}{@{}ccccc@{}}
\toprule
& \multirow{2}{*}{MNIST} & Fashion & \multirow{2}{*}{CIFAR10} & \multirow{2}{*}{CIFAR100} \\
& & MNIST & & \\
\midrule
\standard & $0.92$ & $0.84$ & $0.84$ & $0.66$ \\
\hyperbolic & $0.91$ & $0.81$ & $0.87$ & $0.70$ \\
\spherical & $0.93$ & $0.84$ & $\boldsymbol{0.90}$ & $\boldsymbol{0.84}$ \\
\arcface & $0.95$ & $\boldsymbol{0.89}$ & $\boldsymbol{0.90}$ & $0.80$ \\
\vmf & $\boldsymbol{0.97}$ & $0.88$ & $0.82$ & $0.80$ \\
\bottomrule
\end{tabular}
\end{center}}
\caption{
The mean AUROC indicating how well the norm of an embedding, $\norm{\bz}$, discriminates
correct and incorrect classifier outputs for five losses (rows) and four data sets (columns).
Chance is 0.5; perfect is 1.0. 
Boldface indicates the highest value.
Error bars are negligible across five replications. 
Although the embedding norm signals classifier accuracy for all losses, spherical losses
yield the strongest signal. 
}
\label{tab:auroc_correct}
\end{table}

\subsubsection{von Mises--Fisher Loss}
\label{sec:vmf_loss}

The von Mises--Fisher (vMF) distribution is the maximum-entropy distribution on the surface of a hypersphere, parameterized by a mean unit vector, $\bmu$, and isotropic concentration, $\kappa$.
The pdf for an $n$-dimensional unit vector $\boldsymbol{x}$ is:
\begin{equation}
\begin{split}
    &p(\boldsymbol{x}; \;\bmu, \kappa) = C_n(\kappa) \exp(\kappa \bmu^\text{T} \bx) \text{~with} \\
    &C_n(\kappa) = \frac{\kappa^{n / 2 - 1}}{(2\pi)^{n/2} I_{n/2-1}(\kappa)},
\end{split}
\end{equation}
\noindent where $\bx, \bmu \in \mathbb{S}^{n-1}$, $\kappa \ge 0$, and $I_v$ denotes the modified Bessel function of the first kind at order $v$.

The von Mises--Fisher loss, hereafter \vmf, uses the same form of the posterior as \spherical~(Equation \ref{eq:spherical}), although $\bz$ and $\{ \bw_j \}$
are now vMF random variables, defined in terms of the deterministic output of the network, $\bzt$, and
the learnable weight vector for each class $j$, $\bwt_j$:
\begin{equation}
\begin{split}
    &\bz \sim \text{vMF}\left(\bmu = \frac{\bzt}{\norm{\bzt}}, \kappa = \norm{\bzt}\right), \\
    &\bw_j \sim \text{vMF}\left(\bmu = \frac{\bwt_j}{\norm{\bwt_j}}, \kappa = \norm{\bwt_j}\right).
\end{split}
\end{equation}
The norm $\norm{.}$ directly controls the spread of the distribution with a zero norm yielding a uniform distribution over the hypersphere's surface.
The loss remains the negative log-likelihood under the target class, but in contrast to \spherical,  it is
necessary to marginalize over the the embedding and weight-vector uncertainty:
\begin{equation}
\begin{split}
    \mathcal{L}(y, \bz; \; \bw_{1:Y}) &= \mathbb{E}_{\bz, \bw_{1:Y}}[-\log p(y \vert \bz, \bw_{1:Y})] \\ 
    & = \mathbb{E}_{\bz, \bw_{1:Y}}\left[-\log \frac{\exp(\beta \cos \theta_y))}{\sum_j \exp(\beta \cos \theta_j)} \right],
\end{split}
\end{equation}

\noindent where $Y$ is the total number of classes in the training set. Applying Jensen's inequality, we obtain
an upper-bound on $\mathcal{L}$ which allows us to marginalize over the $\{ \bw_j \}$ and obtain a form expressed in terms of an expectation over $\bz$:
\begin{equation}
\begin{split}
    \mathcal{L}(y&, \bz; \; \bw_{1:Y}) \le \\
    & \mathbb{E}_{\bz}\Bigg[\log \Bigg(\sum_j \exp (\log C_n(\norm{\bwt_j}) \\
    &\hspace{0.31in}- \log C_n (\norm{\bwt_j + \beta \bz}) ) \Bigg) \Bigg] - \beta \mathbb{E}[\bw_y] \mathbb{E}[\bz],
\label{eq:vmf_objective}
\end{split}
\end{equation}
where $\mathbb{E}[\bz] = (I_{n/2}(\kappa) / I_{n/2-1}(\kappa)) \bmu_{\bz}$. 
This objective can be approximated by sampling only from $\bz$ and we find that during both training and testing, 10 samples is sufficient.
At test time, \vmf~approximates $\mathbb{E}_{\bz, \bw_{1:Y}}[p(y \vert \bz, \bw_{1:Y})]$ using Monte Carlo samples from each of $\bz$ and $\{ \bw_j \}$.

To sample, we make use of a rejection-sampling reparameterization trick \cite{Davidson2018}.
However, \cite{Davidson2018} computes modified Bessel functions on the CPU with manually-defined gradients for backpropagation, substantially slowing both the forward and backwards passes through the network.
Instead, we borrow tight bounds for $I_{n/2}(\kappa) / I_{n/2-1}(\kappa)$ from \cite{Ruiz2016} and $\log C_n(\kappa)$ from \cite{Kumar2019vMF}, which together make Equation \ref{eq:vmf_objective} efficient and tractable to compute.
We find that the rejection sampler is stable and efficient in embedding spaces up to at least 512D, adding little overhead to the training and testing procedures (see Appendix \ref{sec:runtime}).
A full derivation of the loss is provided in Appendix \ref{sec:derivation}.
Additional details regarding the bounds for $I_{n/2}(\kappa) / I_{n/2-1}(\kappa)$ and $\log C_n(\kappa)$ can be found in Appendix \ref{sec:bounds}.

The network fails to train when the initial $\{ \bwt_j \}$ are chosen using standard initializers, particularly
in higher dimensional embedding spaces. We discovered the failure to be due to near-zero gradients for
the ratio of modified Bessel functions when the vector norms are small (see flat slope in Figure~\ref{fig:kappa_init} for small $\kappa$). We derived a dimension-equivariant initialization scheme (see Appendix \ref{sec:kappa_init})
that produces norms that yield a strong enough gradient for training. We also include a fixed scale-factor on the embedding, $\bzt$, for the same reason. 
The points in Figure \ref{fig:kappa_init} show the scaling of initial parameters our scheme produces
for various embedding dimensionalities, designed to ensure the ratio of 
modified Bessel functions has a constant value of 0.4.
The expected norms are plotted as individual points with matching color and we find they produce a near-perfect fit for greater than 8D, lying on-top of their corresponding curves.

\begin{figure}[t]
\centering
\includegraphics[width=0.475\textwidth]{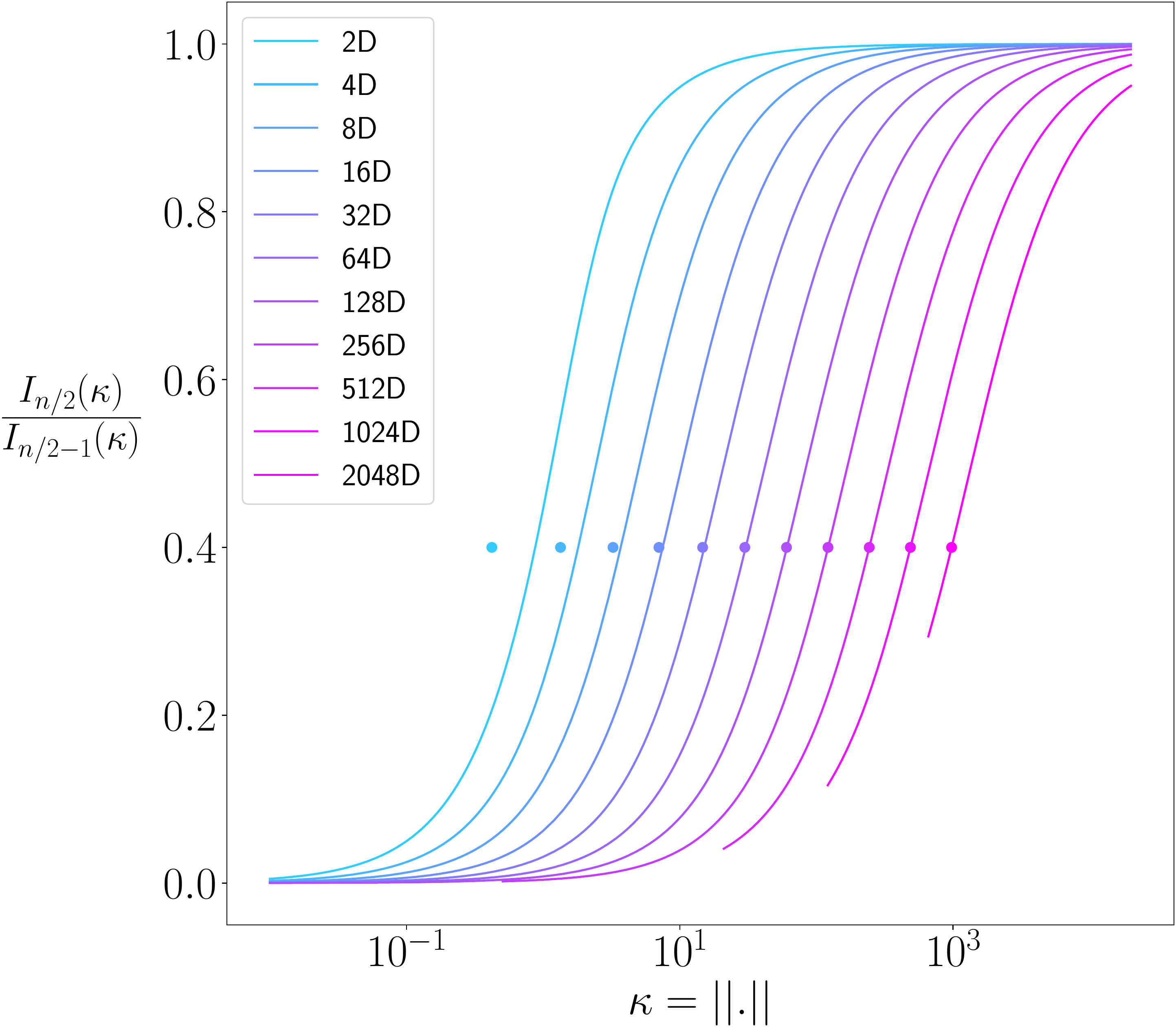}
\caption{The ratio of modified Bessel functions versus $\kappa$ for various embedding dimensionalities (colored curves). Our initializer for $\kappa$ ensures the ratio of modified Bessel functions is constant regardless of the dimensionality. The value of $\kappa$ provided by the initializer for each dimensionality is plotted as a single point. A perfect initializer would ensure the point sits exactly on the matching-colored curve. For this simulation, we initialized such that the y-axis had a constant value of 0.4.}
\label{fig:kappa_init}
\end{figure}

\begin{figure}[t]
\centering
\includegraphics[width=0.471\textwidth]{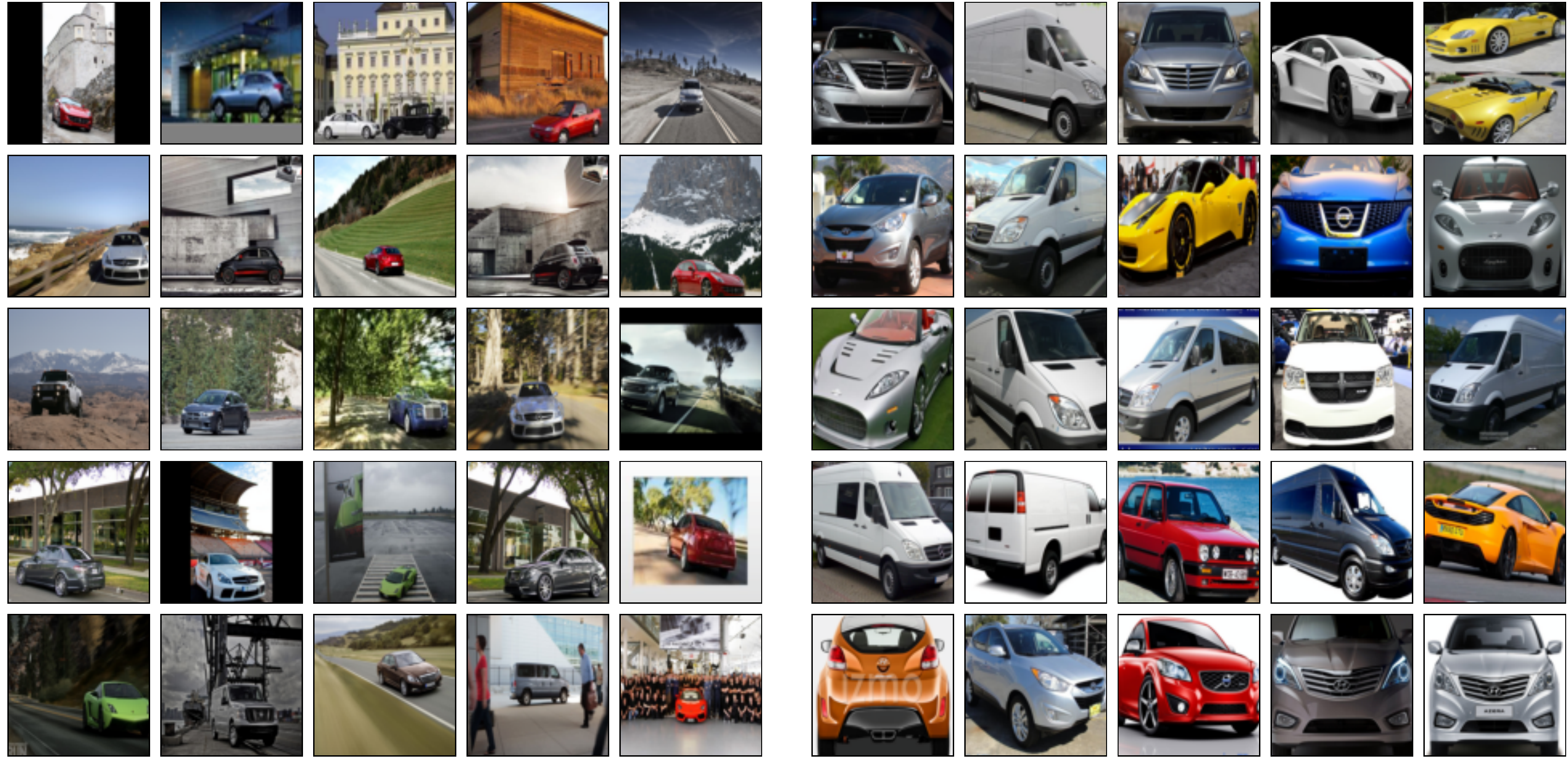}
\caption{Cars196 test images corresponding to \vmf~embeddings with the (left) smallest $\kappa_{\bz}$ and (right) largest $\kappa_{\bz}$. Instances that are more difficult to classify or ambiguous correspond to small $\kappa_{\bz}$.}
\label{fig:vmf_cars}
\end{figure}

To demonstrate that \vmf~learns explicit uncertainty structure, we train it on Cars196 \cite{Krause2013}, a dataset where the class is determined by the make, model, and year of a photographed car. In Figure~\ref{fig:vmf_cars}, we present images corresponding
to embeddings whose distributions have the most uncertainty (smallest $\kappa_z$) and least uncertainty
(largest $\kappa_z$) in the test set.
\vmf~behaves quite sensibly:
the most uncertain embeddings correspond to images of cars that are far from the camera or at poses where it's difficult to extract the make, model, and year; and
the most certain embeddings correspond to images of cars close to the camera in a neutral pose.

\begin{figure*}[t]
\centering
\includegraphics[width=1.0\textwidth]{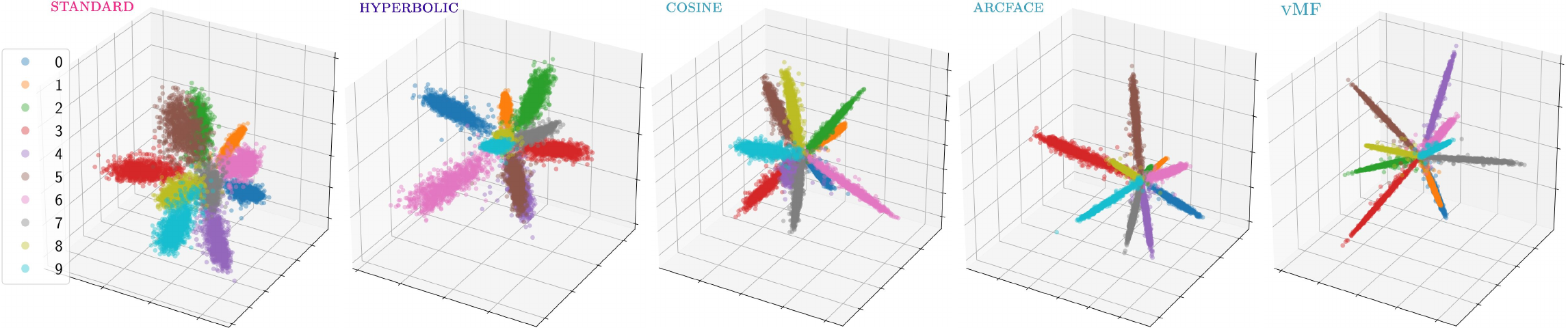}
\caption{3D embeddings of the MNIST test set for each of the five classification variants. Instances are colored by their ground-truth class. The plotted instances for \vmf~correspond to $\bmu_{\bz}$. Note that \hyperbolic, \spherical, \arcface, and \vmf~are showing embeddings prior to the normalization/projection step. Best viewed in color.}
\label{fig:mnist_embeddings}
\end{figure*}

\section{Experimental Results}
We experiment with four fixed-set classification datasets---MNIST \cite{LeCun2010}, FashionMNIST \cite{Xiao2017}, CIFAR10 \cite{Krizhevsky2009}, and CIFAR100 \cite{Krizhevsky2009}---as well as three common datasets for image retrieval---Cars196 \cite{Krause2013}, CUB200-2011 \cite{Wah2011}, and Stanford Online Products (SOP) \cite{Song2016}. 
MNIST and FashionMNIST are trained with 3D embeddings, CIFAR10 and CIFAR100 with 128D embeddings, and all open-set datasets with 512D embeddings.
We perform a hyperparameter search for all losses on each dataset.
The hyperparameters associated with the best performance on the validation set are then used to train five replications of the method.
Reported test performance represents the average over the five replications.
Additional details including the network architecture, dataset details, and hyperparameters are included in Appendix \ref{sec:experimental_details}.

\subsection{Fixed-Set Classification}
\label{sec:fixed_set}

\sidecaptionvpos{table}{c}
\begin{SCtable*}
\begin{tabular}{@{}ccccc@{}}
\toprule
& \multirow{2}{*}{MNIST} & Fashion & \multirow{2}{*}{CIFAR10} & \multirow{2}{*}{CIFAR100} \\
& & MNIST & & \\
\midrule
\standard & $98.92 \pm 0.03$ & $90.31 \pm 0.12$ & $\boldsymbol{94.13 \pm 0.05}$ & $69.21 \pm 0.18$ \\
\hyperbolic & $98.92 \pm 0.03$ & $90.31 \pm 0.13$ & $\boldsymbol{94.11 \pm 0.05}$ & $69.85 \pm 0.07$ \\
\spherical & $98.99 \pm  0.03$ & $90.39 \pm 0.09$ & $93.99 \pm 0.10$ & $\boldsymbol{70.57 \pm 0.54}$ \\
\arcface & $\boldsymbol{99.13 \pm 0.02}$ & $\boldsymbol{90.73 \pm 0.12}$ & $\boldsymbol{94.15 \pm 0.05}$ & $69.08 \pm 0.57$ \\
\vmf & $99.02 \pm 0.04$ & $\boldsymbol{90.82 \pm 0.14}$ & $\boldsymbol{94.00 \pm 0.12}$ & $\boldsymbol{69.94 \pm 0.18}$ \\
\bottomrule
\end{tabular}
\caption{Mean classification accuracy (\%) of each loss across four fixed-set classification tasks. Error bars represent $\pm 1$ standard-error of the mean. Boldface indicates the best-performing loss(es). Note that on average, the three spherical losses outperform \hyperbolic\ and \standard.}
\label{tab:fixed_set_acc}
\end{SCtable*}

\begin{table*}
\begin{center}
\begin{tabular}{@{}ccccccccc@{}}
\toprule
& \multicolumn{4}{c}{ECE} & \multicolumn{4}{c}{ECE with Temperature Scaling} \\
\cmidrule(lr){2-5}
\cmidrule(lr){6-9}
& \multirow{2}{*}{MNIST} & Fashion & \multirow{2}{*}{CIFAR10} & \multirow{2}{*}{CIFAR100} & \multirow{2}{*}{MNIST} & Fashion & \multirow{2}{*}{CIFAR10} & \multirow{2}{*}{CIFAR100} \\
& & MNIST & & & & MNIST & & \\
\midrule
\standard & $2.4 \pm 0.2$ & $12.4 \pm 0.8$ & $8.8 \pm 0.1$ & $20.6 \pm 0.2$ & $\boldsymbol{0.4 \pm 0.1}$ & $\boldsymbol{5.5 \pm 1.3}$ & $\boldsymbol{2.7 \pm 0.2}$ & $\boldsymbol{2.2 \pm 0.1}$ \\
\hyperbolic & $2.8 \pm 0.1$ & $13.2 \pm 0.4$ & $8.9 \pm 0.1$ & $22.0 \pm 0.2$ & $1.4 \pm 0.1$ & $7.2 \pm 0.7$ & $\boldsymbol{2.8 \pm 0.1}$ & $2.6 \pm 0.1$ \\
\spherical & $\boldsymbol{1.8 \pm  0.2}$ & $7.9 \pm 0.5$ & $8.9 \pm 0.2$ & $21.8 \pm 1.2$ & $1.6 \pm  0.1$ & $\boldsymbol{4.2 \pm 0.1}$ & $6.4 \pm 0.2$ & $10.8 \pm 0.8$ \\
\arcface & $2.3 \pm 0.1$ & $11.0 \pm 0.5$ & $10.0 \pm 0.1$ & $26.4 \pm 0.4$ & $1.7 \pm 0.1$ & $7.2 \pm 0.4$ & $8.7 \pm 0.1$ & $15.7 \pm 0.2$ \\
\vmf & $\boldsymbol{1.6 \pm 0.1}$ & $\boldsymbol{4.2 \pm 0.5}$ & $\boldsymbol{5.9 \pm 0.2}$ & $\boldsymbol{7.9 \pm 0.3}$ & $1.5 \pm 0.1$ & $5.0 \pm 0.2$ & $5.3 \pm 0.2$ & $8.0 \pm 0.2$ \\
\bottomrule
\end{tabular}
\end{center}
\caption{Mean expected calibration error (\%), computed with 15 equal-mass bins, before post-hoc calibration (leftmost four columns) and after temperature scaling (rightmost four columns) across the four fixed-set classification tasks. Error bars represent $\pm 1$ standard-error of the mean. Boldface indicates the loss(es) with the lowest error.}
\label{tab:fixed_set_ece}
\end{table*}

We begin by comparing representations learned by the five losses we described:
\standard, \hyperbolic, \spherical, \arcface, and \vmf.  The latter three have spherical geometries. \arcface\ is
a minor variant of \spherical\ claimed to be a top performer for face verification \cite{Deng2019}. \vmf\ is our
probabilistic extension of \spherical.
Using a 3D embedding on MNIST, we observe decreasing intra-class angular variance for the losses 
appearing from left to right in Figure~\ref{fig:mnist_embeddings}. The intra-class variance is related to inter-class discriminability, as the 10 classes are similarly dispersed for all losses.
The three losses with spherical geometry obtain the lowest variance, with
\arcface\ lower than \spherical\ due to a margin hyperparameter designed to penalize intra-class variance; and
\vmf\ achieves the same, if not lower variance still, as a natural consequence of uncertainty reduction.

The test accuracy for each of the fixed-set classification datasets and the five losses is presented in Table \ref{tab:fixed_set_acc}. 
Across all datasets, spherical losses outperform \standard~and \hyperbolic.
Among the spherical losses, 
\arcface\ and \spherical\ are deficient on at least one dataset, whereas
\vmf\ is a consistently strong performer.

Table \ref{tab:fixed_set_ece} presents the top-label 
expected calibration error (ECE) for each dataset and
loss. 
The top-label expected calibration error approximates the disparity between a model's confidence output, $\max_y p(y \vert \bz)$, and the ground-truth likelihood of being correct \cite{Kumar2019Calibration,Roelofs2021}.
The left four columns show out-of-the-box ECE on the test set (i.e., prior to any post-hoc calibration).
\vmf~has significantly reduced ECE compared to other losses, with relative error reductions of 40-70\% for FashionMNIST, CIFAR10, and CIFAR100.
The right four columns show ECE after applying post-hoc temperature scaling
\cite{Guo2017}.
\standard~and \hyperbolic~greatly benefit from temperature scaling, with \standard~exhibiting the lowest calibration error.

Post-hoc calibration requires a validation set, but many settings cannot
afford the data budget to reserve a sizeable validation set, which makes
out-of-the-box calibration a desirable property. For example,
in few-shot and transfer learning, one may not have enough data in the target
domain to both fine tune the classifier and validate.

Temperature scaling is not as effective when applied to spherical losses as when
applied to \standard\ and \hyperbolic. The explanation, we hypothesize, is that 
the spherical losses incorporate a learned temperature parameter $\beta$ (which
we discuss below), which is unraveled by the calibration temperature.
We leave it as an open question for how to properly post-hoc calibrate spherical losses.

\subsection{Open-Set Retrieval}
\label{sec:open_set_retrieval}

\sidecaptionvpos{table}{c}
\begin{SCtable*}
\begin{tabular}{@{}ccccc@{}}
\toprule
& Cars196 & CUB200-2011 & SOP \\
\midrule
\standard & $21.3 \pm 0.2$ & $20.0 \pm 0.2$ & $39.7 \pm 0.1$ \\
\textcolor{standard}{\textsc{+ cosine at test}} & $23.2 \pm 0.1$ & $21.4 \pm 0.2$ & $42.1 \pm 0.1$ \\
\hyperbolic & $22.9 \pm 0.3$ & $20.1 \pm 0.3$ & $41.0 \pm 0.2$ \\
\textcolor{hyperbolic}{\textsc{+ cosine at test}} & $25.0 \pm 0.4$ & $21.8 \pm 0.2$ & $44.0 \pm 0.1$ \\
\spherical & $24.6 \pm 0.4$ & $\boldsymbol{22.8 \pm 0.1}$ & $\boldsymbol{44.3 \pm 0.1}$ \\
\arcface & $\boldsymbol{27.4 \pm 0.2}$ & $\boldsymbol{23.1 \pm 0.3}$ & $40.8 \pm 0.3$ \\
\vmf & $\boldsymbol{27.2 \pm 0.1}$ & $22.1 \pm 0.1$ & $38.3 \pm 0.2$  \\
\bottomrule
\end{tabular}
\caption{Mean mAP@R (\%) across the three open-set image retrieval tasks. Error bars represent $\pm 1$ standard-error of the mean. Boldface indicates the best-performing loss(es). ``+ Cosine at Test'' replaces the default metric for the geometry (i.e., Euclidean for \standard~and Poincar\'e for \hyperbolic) with cosine distance to compare instances.}
\label{tab:open_set_acc}
\end{SCtable*}

For open-set retrieval, we follow the data preprocessing pipeline of \cite{Boudiaf2020} where each dataset is first split into a train set and test set with disjoint classes.
We additionally split off 15\% of the training classes for a validation set, a decision that has been left out of many training procedures of similarity-based losses \cite{Musgrave2020}.
We evaluate methods using mean-average-precision at R (mAP@R), a metric shown to be more informative than Recall@1 \cite{Musgrave2020}.
For the stochastic loss, \vmf, we compute $\mathbb{E}_{\bz_{1:N}}[\text{mAP@R}(\bz_{1:N}, y_{1:N})]$, where $N$ is the number of test instances.

Table \ref{tab:open_set_acc} presents the retrieval performance for each loss.
As with fixed-set classification, there is no consistent winner across all datasets, but spherical losses tend to outperform \standard~and \hyperbolic.
Boudiaf \etal~\cite{Boudiaf2020} find that retrieval performance can be improved for \standard~by employing cosine distance at test time.
Although no principled explanation is provided, we note from Figure~\ref{fig:mnist_embeddings} that
$\norm{\bz}$ introduces a large source of intra-class variance in
\standard~and \hyperbolic.
This variance is factored out automatically by the spherical losses.
As shown in Table~\ref{tab:open_set_acc}, cosine distance at test improves both
\standard\ and \hyperbolic, though not to the level of the best performing spherical loss.

In contrast to other stochastic losses \cite{Scott2019,Oh2019}, \vmf~scales to high-dimensional embeddings---512D in this case---and can be competitive with state-of-the-art.
However, it has the worst performance on SOP which has 9,620 training classes; many more than all other datasets.
In Section \ref{sec:vmf_loss}, we mentioned that to marginalize out the weight distributions, we successively apply Jensen's inequality to their expectations.
The decision resulted in a tractable loss, but it is an upper-bound on the true loss and the bound very likely becomes loose as the number of training classes increases.
We hypothesize the inferior performance is due to this design choice, and despite experimentation with curriculum learning techniques to condition on a subset of classes during training, results did not improve.

\subsection{Role of Temperature}

Due to spherical losses using cosine similarity, the logits are bounded in $[-1, 1]$.
Consequently, it is necessary to scale the logits to a more suitable range. 
In past work, spherical losses have incorporated an inverse-temperature constant,
$\beta > 0$ \cite{Zhai2019,Deng2019,Wang2018,Wang2018a}.
Past efforts to turn $\beta$ into a trainable parameter find that it either does not
work as well as fixing it, or no comparison is made to a fixed value
\cite{Ranjan2017,Ranjan2019,Wang2017normface}.

\begin{figure}[t]
\centering
\includegraphics[width=0.475\textwidth]{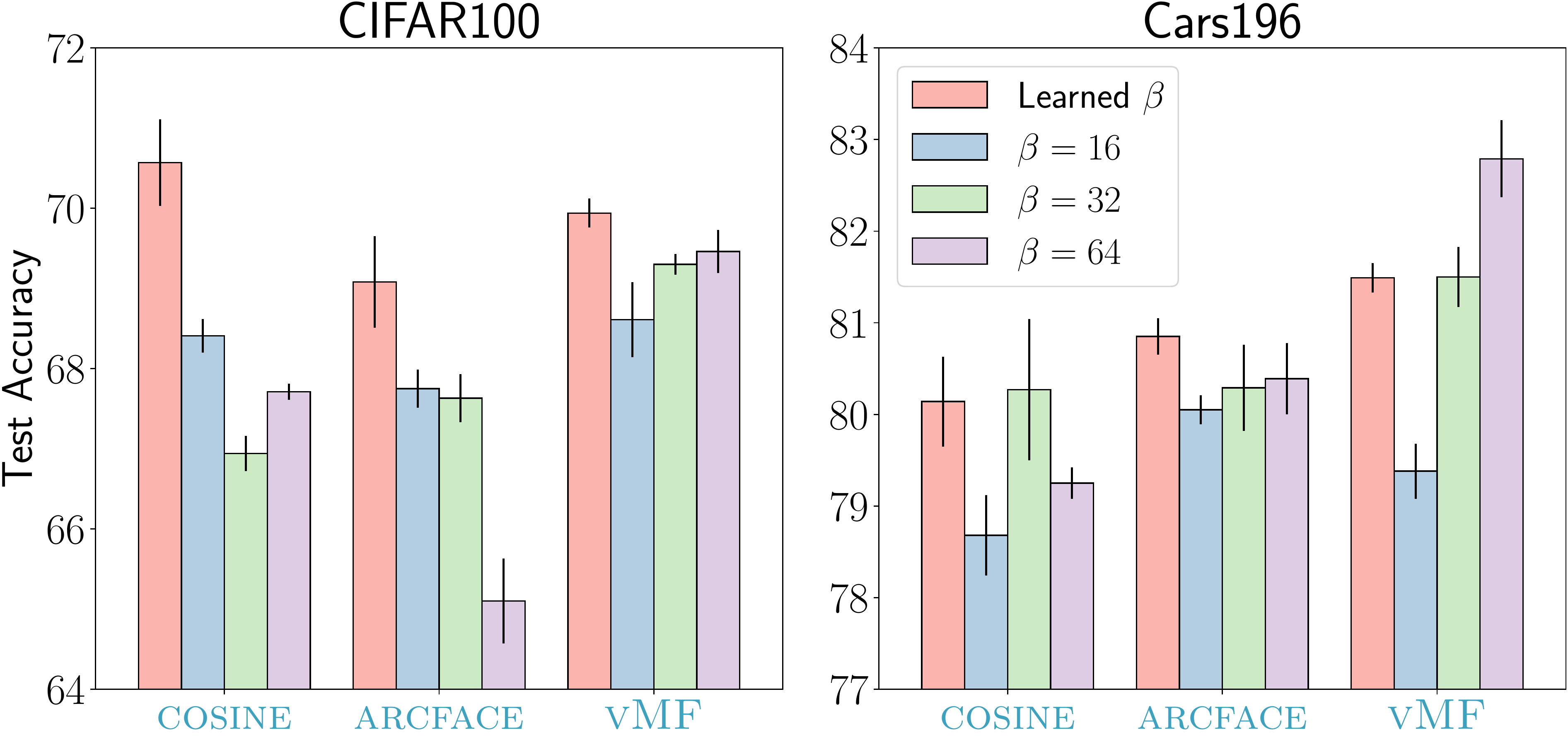}
\caption{Comparison between a learned temperature and various values of a fixed temperature on (left) CIFAR100 and (right) Cars196. Learning the temperature performs at least as well as fixing it, with exception to \vmf~on Cars196.}
\label{fig:fixed_temp}
\end{figure}

In \spherical, \arcface, and \vmf, we compare a fixed $\beta$ to a trained 
$\beta$, using fixed values common in the literature. Under a suitable initialization
and parameterization of temperature, the trained $\beta$ performs at least as well 
as a fixed value and avoids the manual search (Figure \ref{fig:fixed_temp}).
In particular, rather than performing constrained optimization 
in $\beta$, we perform unconstrained optimization in $\tau = \log(\beta)$.
Details can be found in Appendix~\ref{sec:temp_details}.

\section{Related Work}

In Section \ref{sec:intro}, we dichotomized the literature in a coarse manner
based on whether losses are similarity based or classification based.
In this section, we focus on recent work relevant to the \vmf, including losses functions using vMF distributions and stochastic classifiers.

A popular use of vMF distributions in machine learning has been clustering \cite{Banerjee2005,Gopal2014}.
Banerjee \etal~\cite{Banerjee2005} consider a vMF mixture model trained with expectation-maximization, and Gopal and Yang \cite{Gopal2014} propose a fully-Bayesian extension along with hierarchical and temporal versions trained with variational inference.
In addition, vMF distributions have begun being applied in supervised settings.
Hasnat \etal~\cite{Hasnat2017} suggest a supervised classifier for face verification where each term in the softmax is interpreted as the probability density of a vMF distribution.
Zhe \etal~\cite{Zhe2018} propose a similarity-based loss that is functionally identical to \cite{Hasnat2017}, except the mean parameters of the vMF distributions for each class are estimated as the maximum-likelihood estimate of the training data.
Park \etal~\cite{Park2019} describe the spherical analog to the prototypical network \cite{Snell2017}, but use a generator network to output the prototypes.
The downside of these supervised losses compared to \vmf~is they assume the concentration parameter across all classes is identical and fixed, canceling $C_n(\kappa)$ in the softmax.
Such a decision is mathematically convenient, but removes a significant amount of flexibility in the model.
Davidson \etal~\cite{Davidson2018} propose a variational autoencoder (VAE) with hyperspherical latent structure by using a vMF distribution. 
They find it can outperform a Gaussian VAE, but only in lower-dimensional latent spaces.
We leverage the rejection-sampling reparameterization scheme they compose to train \vmf. 

Other work has sought losses that consider either stochastic embeddings or stochastic logits, but they suggest Gaussians instead of vMFs.
Chang \etal~\cite{Chang2020} suppose stochastic embeddings, but deterministic classification weights, and train a classifier using Monte Carlo samples. They also add a KL-divergence regularizer between the embedding distribution and a zero-mean, unit-variance Gaussian.
Collier \etal~\cite{Collier2020} propose a classification loss with stochastic logits that uses a temperature-parameterized softmax. They show it can train under the influence of heteroscedastic label noise with improved accuracy and calibration.
Shi and Jain \cite{Shi2019} convert deterministic face embeddings into Gaussians by training a post-hoc network to estimate the covariances. Their objective maximizes the mutual likelihood of same-class embeddings.
Scott \etal~\cite{Scott2019} and Oh \etal~\cite{Oh2019} propose similarity-based losses that are the stochastic analogs of the prototypical network \cite{Snell2017} and pairwise contrastive loss \cite{Hadsell2006}, respectively.
Neither loss shows promise with high-dimensional embeddings. The loss from \cite{Scott2019} struggles to compete with a standard prototypical network in 64D, and \cite{Oh2019} omits any results with embeddings larger than 3D.
Gaussian distributions suffer from the curse of dimensionality \cite{Davidson2018}; one possible explanation for the inferior performance compared to deterministic alternatives.

The work most closely related to ours is Kornblith \etal~\cite{Kornblith2020}, 
which compares \standard, \spherical, and an assortment of alternatives including
mean-squared error, sigmoid cross-entropy, and various regularizers applied to
\standard. In contrast, we focus on multiple variants of spherical losses and 
comparing geometries. Their findings are compatible with ours.

\section{Conclusions}

In this work, we perform a systematic comparison of classification losses that span three embedding geometries---Euclidean, hyperbolic, and spherical---and attempt to reconcile the discrepancies in past work regarding their performance.
Our investigations have led to a stochastic spherical classifier where embeddings and class weight vectors are von Mises--Fisher random variables.
Our proposed loss is on par with other classification variants and also produces consistently reduced out-of-the-box calibration error. Our loss encodes
instance ambiguity using the concentration parameter of the vMF distribution.

Consistent with the no-free-lunch theorem \cite{Wolpert1997}, we find there is no one
loss to rule them all. Scanning arXiv, this conclusion is uncommon, and it is even
rarer in published work.
The performance jump claimed for novel losses often vanishes with rigorous,
systematic comparsions in controlled settings---in
settings where the network architecture is identical, hyperparameters are optimized with equal vigor, regularization and data augmentation is matched, and a held-out validation set is used to choose hyperparameters.
Musgrave \etal~\cite{Musgrave2020} reach a similar conclusion: the gap between 
various similarity-based losses, as well as the spherical classification 
losses (\spherical~and \arcface) was much smaller in magnitude than was 
claimed in the original papers proposing them.
We are hopeful that future work on supervised loss functions will also prioritize rigorous experiments and step away from the compulsion to show unqualified
improvements over state of the art.

Pedantics aside, we are able to glean some positive messages by systematically
comparing performance across geometries and across different classification paradigms.
We focus on specific recommendations in the remainder of the paper that address trade-offs among the embedding geometries.

\paragraph{Accuracy.} Many losses are designed primarily with accuracy in mind.
Across both fixed- and open-set tasks, we find that losses operating on a 
spherical geometry perform best. Our results support
the following ranking of losses: \standard~$\le$ \hyperbolic~$\le$ \{\spherical, \arcface, \vmf\}.
Additionally, our results corroborate two conclusions from Chen \etal~\cite{Chen2020}.
First, smaller intra-class variance yields better generalization, and second, \standard\ focuses too much on separating classes by increasing embedding norms rather than reducing angular variance (e.g.,~Figure~\ref{fig:mnist_embeddings}).
Although the best of the spherical losses appears to be dataset dependent,
the guidance to focus on spherical losses and perform empirical comparisons
is not business as usual for practitioners, who treat \standard\ as the go-to loss.
For practitioners using models with non-spherical geometries 
(\standard\ and \hyperbolic), we can still provide the guidance to use cosine
distance at test time---discarding the embedding magnitude---which seems to
reliably lead to improved retrieval performance.

An aspect of accuracy we do not consider, however, is the performance of downstream target tasks where a classification loss is used to pre-train weights.
Kornblith \etal~\cite{Kornblith2020} discover that better class separation on the pre-training task can lead to worse transfer performance, as improved discriminability implies task specialization (i.e., throwing away inter-class variance necessary to relate classes).
Spherical losses perform well on fixed-set tasks as well as on open-set tasks where the novel test classes are drawn from a distribution very similar to that of the
training distribution, but transfer learning is typically a setting where 
\standard~has been shown to be superior.
We believe that it would be useful in future research to distinguish between
near- and far-transfer, as doing so may yield distinct conclusions.

\paragraph{Calibration.} In sensitive domains, deployed machine learning systems 
must produce trustworthy predictions, which requires that model confidence scores
match model accuracy (i.e., they must be well calibrated).
To our knowledge, we are the first to rigorously examine the effect of
embedding geometry on calibration performance.
Our findings indicate that when a validation set is available for temperature scaling, \standard~consistently produces predictions with the lowest calibration error, but 
\standard~generally underperforms in accuracy.
Additionally, there does not exist a significant gap in out-of-the-box calibration performance across previously proposed losses from the three geometries.
However, our novel \vmf~loss achieves superior calibration while maintaining state-of-the-art classification performance.

\paragraph{Future Work.} We are exploring several directions for future work.
First, \cite{Decao2020} introduces a novel probability distribution---the Power Spherical distribution---with support on the surface of the hypersphere.
They claim it has several key improvements over vMF distributions including a reparameterization trick that does not require rejection sampling, improved stability in high dimensions and for large values of the concentration parameter, and no dependence on Bessel functions.
A stochastic classifier based on the Power Spherical distribution would likely improve the computational efficiency as well as the optimization procedure, particularly for high-dimensional embedding spaces.
Second, we note that our formulation of \vmf~is a special case of a class of objectives based on the deep variational information bottleneck \cite{Alemi2016}: $I(Z, Y) - \gamma I(Z, X)$, where $\gamma = 0$ and the classification weights are also stochastic.
Our objective thus lacks a regularizer attempting to compress the amount of input information contained in the embedding.
Adding this regularization term may lead to improved performance and robustness.
Third, all of the experimental datasets are approximately balanced and without label noise.
Due to the stochasticity of the classification weights, \vmf~seems likely to benefit 
in supervised settings with long-tailed distributions or heteroscedastic label noise \cite{Collier2020}.

It is unfortunate that the field of supervised representation learning has become so vast that researchers tend to specialize
in a particular learning paradigm (e.g., fixed-set classification, few-shot
learning, transfer learning, deep metric-learning) or domain (e.g., face verification,
person re-identification, image classification). As a result, losses are often
pigeonholed to one paradigm or domain. The objective of this work is to lay out 
the space of losses in terms of embedding geometry and systematically survey 
losses that are not typically compared to one another. One surprising and important
result in this survey is the strength of spherical losses, and the resulting
dissociation of embedding norm and output confidence.

\clearpage

{\small
\bibliographystyle{ieee_fullname}
\bibliography{main}
}

\clearpage

\onecolumn{
\begin{center}
    \textbf{\LARGE Supplementary Material}
\end{center}
\vspace{0.15in}
\appendix
\section{Derivation of the von Mises--Fisher Loss}
\label{sec:derivation}

Let $\bz$ and $\{\bw\}_{1:Y}$ be von Mises--Fisher random variables for the embedding and class weight vectors, respectively, and $Y$ be the number of training classes.
In addition, let $\bwt_j$ be the learnable weight vector for class $j$ that is used to parameterize $\bw_j$.

\begin{align*}
    \mathcal{L}(y, &\bz; \; \bw_{1:Y}) = \mathbb{E}_{\bz, \bw_{1:Y}}[-\log p(y \vert \bz, \bw_{1:Y})] \notag \\ 
    & = \mathbb{E}_{\bz, \bw_{1:Y}}\left[-\log \frac{\exp \left(\beta \cos \theta_y \right)}{\sum_j \exp(\beta \cos \theta_j)} \right] \notag \\
    & = \mathbb{E}_{\bz, \bw_{1:Y}}\left[\log \left(\sum_j \exp \left(\beta \bw_j^{\text{T}} \bz \right) \right) \right] - \beta \mathbb{E}[\bw_y] \mathbb{E}[\bz] \tag*{(Independence of $\bz$ and $\bw_y$)} \\
    & \le \mathbb{E}_{\bz}\left[\log \left(\sum_j \mathbb{E}_{\bw_j} \left[\exp \left(\beta \bw_j^{\text{T}} \bz \right)\right] \right) \right] - \beta \mathbb{E}[\bw_y] \mathbb{E}[\bz] \tag*{(Jensen's inequality with respect to $\{\bw_j\}$)} \\
    & = \mathbb{E}_{\bz}\left[\log \left(\sum_j \int_{\bw_j} C_n(\kappa_{\bw_j}) \exp(\kappa_{\bw_j} \bmu_{\bw_j}^{\text{T}} \bw_j) \exp(\beta \bw_j^{\text{T}} \bz) \; \text{d}\bw_j \right)\right] - \beta \mathbb{E}[\bw_y] \mathbb{E}[\bz] \notag \\
    & = \mathbb{E}_{\bz}\left[\log \left(\sum_j C_n(\kappa_{\bw_j}) \int_{\bw_j} \exp((\kappa_{\bw_j} \bmu_{\bw_j} + \beta \bz)^{\text{T}} \bw_j) \; \text{d}\bw_j \right)\right] - \beta \mathbb{E}[\bw_y] \mathbb{E}[\bz] \notag \\
    & = \mathbb{E}_{\bz}\left[\log \left(\sum_j C_n(\kappa_{\bw_j}) \int_{\bw_j} \exp \left(\frac{\norm{\kappa_{\bw_j} \bmu_{\bw_j} + \beta \bz}}{\norm{\kappa_{\bw_j} \bmu_{\bw_j} + \beta \bz}} (\kappa_{\bw_j} \bmu_{\bw_j} + \beta \bz)^{\text{T}} \bw_j \right) \; \text{d}\bw_j \right)\right] - \beta \mathbb{E}[\bw_y] \mathbb{E}[\bz] \notag \\
    & = \mathbb{E}_{\bz}\left[\log \left(\sum_j \frac{C_n(\kappa_{\bw_j})}{C_n \left(\norm{\kappa_{\bw_j} \bmu_{\bw_j} + \beta \bz}\right)} \int_{\bw_j} \text{vMF}\left(\bw_j; \; \bmu = \frac{\kappa_{\bw_j} \bmu_{\bw_j} + \beta \bz}{\norm{\kappa_{\bw_j} \bmu_{\bw_j} + \beta \bz}}, \kappa = \norm{\kappa_{\bw_j} \bmu_{\bw_j} + \beta \bz} \right) \; \text{d}\bw_j \right)\right] \notag \\
    & \hspace{5.7in} - \beta \mathbb{E}[\bw_y] \mathbb{E}[\bz] \notag \\
    & = \mathbb{E}_{\bz}\left[\log \left(\sum_j \frac{C_n(\norm{\bwt_j})}{C_n \left(\norm{\bwt_j + \beta \bz}\right)} \right)\right] - \beta \mathbb{E}[\bw_y] \mathbb{E}[\bz]. \notag \\
\end{align*}

\section{Bounds for $I_{n/2}(\kappa) / I_{n/2-1}(\kappa)$ and $\log C_n(\kappa)$}
\label{sec:bounds}

To approximate $I_{n/2}(\kappa) / I_{n/2-1}(\kappa)$ where $n$ is the dimensionality of the embedding space, we borrow a lower bound from Theorem 4 and an upper bound from Theorem 2 of \cite{Ruiz2016}:

\begin{equation}
\label{eq:bounds}
    \frac{\kappa}{\frac{n-1}{2} + \sqrt{\left(\frac{n+1}{2}\right)^2 + \kappa^2}} \le \frac{I_{n/2}(\kappa)}{I_{n/2-1}(\kappa)} \le \frac{\kappa}{\frac{n-1}{2} + \sqrt{\left(\frac{n-1}{2}\right)^2 + \kappa^2}}.
\end{equation}
Let's denote the lower bound as $g_n(\kappa)$ and the upper bound as $h_n(\kappa)$. Our approximation for the ratio of modified Bessel functions is thus:
\begin{equation}
\label{eq:bessel_approx}
    \frac{I_{n/2}(\kappa)}{I_{n/2-1}(\kappa)} \approx \frac{1}{2}(g_n(\kappa) + h_n(\kappa)).
\end{equation}

Next, we borrow a clever trick from \cite{Kumar2019vMF}.
Note that:
\begin{equation}
    \frac{\text{d}}{\text{d}\kappa}\log C_n(\kappa) = -\frac{I_{n/2}(\kappa)}{I_{n/2-1}(\kappa)}.
\end{equation}
If we plug in our approximation for the ratio of modified Bessel functions from Equation \ref{eq:bessel_approx} and integrate with respect to $\kappa$, we arrive at the following approximation:
\begin{equation}
\begin{split}
    \log C_n(\kappa) &\approx \frac{n-1}{4} \log \left(\frac{n-1}{2} + \sqrt{\left(\frac{n-1}{2}\right)^2 + \kappa^2} \right) - \frac{1}{2} \sqrt{\left(\frac{n-1}{2}\right)^2 + \kappa^2} \\
    & + \frac{n-1}{4} \log \left(\frac{n-1}{2} + \sqrt{\left(\frac{n+1}{2}\right)^2 + \kappa^2} \right) - \frac{1}{2} \sqrt{\left(\frac{n+1}{2}\right)^2 + \kappa^2} + \eta,
\end{split}
\end{equation}
where $\eta$ is an unknown constant resulting from indefinite integration.
While the approximation is messy, it is easy to compute and backpropagate through on accelerated hardware.
We can rewrite the first term of the final objective using an exponential-log trick that allows us to apply the approximation of $\log C_n(\kappa)$ and cancel $\eta$:
\begin{equation}
\label{eq:final_objective}
\begin{split}
    \mathcal{L}(y, \bz; \; \bw_{1:Y}) &\le \mathbb{E}_{\bz}\left[\log \left(\sum_j \frac{C_n(\norm{\bwt_j})}{C_n \left(\norm{\bwt_j + \beta \bz}\right)} \right)\right] - \beta \mathbb{E}[\bw_y] \mathbb{E}[\bz] \\
    &= \mathbb{E}_{\bz}\left[\log \left(\sum_j \exp \left(\log C_n(\norm{\bwt_j}) - \log C_n \left(\norm{\bwt_j + \beta \bz}\right) \right) \right)\right] - \beta \mathbb{E}[\bw_y] \mathbb{E}[\bz].
\end{split}
\end{equation}
Equation \ref{eq:final_objective} is the final form of the \vmf~objective.

\section{Initialization of the Concentration for vMF}
\label{sec:kappa_init}

We seek an initialization of $\kappa$ for all vMF distributions such that the ratio of modified Bessel functions is a constant greater than zero. Such an initialization would ensure gradients are strong enough for training (see Figure \ref{fig:kappa_init}) and all vMF distributions are approximately equally concentrated.
If we take the upper bound, $h_n(\kappa)$, from Equation \ref{eq:bounds} and set it equal to a constant, $\lambda$, and solve for $\kappa$, we get:
\begin{equation}
\label{eq:const_kappa}
    \kappa = \frac{\lambda}{1 - \lambda^2} (n - 1).
\end{equation}

For initializing the concentration of the embedding distribution, $\kappa_{\bz}$, we introduce a constant scalar, denoted $\alpha$, that is multiplied with the embedding output of the network, $\bzt$, prior to $\ell_2$-normalization. The multiplier does not add any flexibility to the network and also does not affect the direction of the embedding. Since $\kappa_{\bz} = \norm{\bzt}$, we estimate the value of $\alpha$ such that the expected embedding norm over the training dataset is the value we desire from Equation \ref{eq:const_kappa}:
\begin{equation}
\begin{split}
    &\mathbb{E}_{\bzt}\left[\norm{\alpha \bzt} \right] = \frac{\lambda}{1 - \lambda^2} (n - 1) \\
    & \Rightarrow \mathbb{E}_{\bzt}\left[\sqrt{\sum_i (\alpha \bzt_i)^2} \right] = \frac{\lambda}{1 - \lambda^2} (n - 1),
\end{split}
\end{equation}
where $\bzt_i$ is the $i$th element of $\bzt$.
Under the strong assumption that $\bzt_1^2 = \bzt_2^2 = \dots = \bzt_n^2$, we have:
\begin{equation}
\begin{split}
\label{eq:alpha}
    &\mathbb{E}_{\bzt_i}\left[\sqrt{n (\alpha \bzt_i)^2} \right] = \frac{\lambda}{1 - \lambda^2} (n - 1) \\
    & \Rightarrow \alpha = \frac{\lambda (n - 1)}{(1 - \lambda^2) \sqrt{n} \; \mathbb{E}[\vert \bzt_i \vert]}.
\end{split}
\end{equation}
Prior to any training, we compute $\mathbb{E}_{\bzt_i}[\vert \bzt_i \vert]$ by passing all of the training data through the network and taking the mean of the resulting tensor across both the embedding-dimension axis and batch axis, prior to $\ell_2$-normalization.
Then, we are able to determine $\alpha$, which is fixed for the duration of training.

We also need to initialize the class weight vectors, $\{ \bwt_j \}$, such that their expected $\ell_2$-norm matches that of $\alpha \bzt$.
Let's denote $\bwt_{j,i}$ as the $i$th element of the weight vector for class $j$. Assume each element, $\bwt_{j,i} \; \forall \; i = 1 \dots n$ is independently and identically distributed according to a Gaussian with zero mean and unknown standard deviation. Borrowing Equation \ref{eq:alpha}:
\begin{equation}
    \xi \mathbb{E}[\vert \bwt_{j,i} \vert] = \frac{\lambda (n - 1)}{(1 - \lambda^2) \sqrt{n}}.
\end{equation}
Note that $\mathbb{E}[\vert \bwt_{j,i} \vert] = \sqrt{\frac{2}{\pi}} \sigma$ for a zero-mean Gaussian. Thus:
\begin{equation}
    \sigma = \frac{\lambda (n - 1)}{(1 - \lambda^2) \sqrt{n}} \sqrt{\frac{\pi}{2}} \frac{1}{\xi}.
\end{equation}
We empirically determined that $\xi = \sqrt{\frac{\pi}{2}}$ produced a near-perfect fit for 8D embedding spaces and larger (see Figure \ref{fig:kappa_init}) resulting in:
\begin{equation}
\bwt_{j,i} \sim \mathcal{N}\left(\mu = 0, \sigma = \frac{\lambda (n - 1)}{(1 - \lambda^2) \sqrt{n}} \right) \; \forall \; j = 1 \dots Y \; \text{and} \; i = 1 \dots n.
\end{equation}
For training \vmf, $\lambda$ is a hyperparameter. See Section \ref{sec:hyper} for the values we used.

\section{Experimental Details}
\label{sec:experimental_details}

\subsection{Architectures}

\paragraph{MNIST \& FashionMNIST.} Experiments on MNIST and FashionMNIST use an architecture adapted from \cite{Oh2019}.
The architecture has two convolutional blocks each with a convolutional layer followed by batch normalization, ReLU, and $2 \times 2$ max-pooling.
The first convolutional layer has a $5 \times 5$ kernel, 6 filters, zero-padding of length 2, and a stride of 1.
The second convolutional layer is identical to the first, but with 16 filters.
After the second convolutional block, the representation is flattened and passed through a fully-connected layer with 120 units, followed by batch normalization and a ReLU.
The representation is finally passed through two fully-connected layers, the first from 120 units to $n$ units, where $n$ is the dimensionality of the embedding space, and then from $n$ units to $Y$ units, where $Y$ is the total number of classes in the training set.
The last fully-connected layer has no bias parameters.
All biases in the network are initialized to zero and all weights are initialized with Xavier uniform.
For \vmf, instead of using Xavier uniform, the weights in the final fully-connected classification layer parameterize vMF distributiuons for each class and thus are initialized using the scheme detailed in Appendix \ref{sec:kappa_init}.
For both MNIST and FashionMNIST, $n = 3$ and $Y = 10$.

\paragraph{CIFARs \& Open-Set.} Experiments on CIFAR10, CIFAR100, Cars196, CUB200-2011, and SOP use a ResNet50 \cite{He2016}.
For CIFAR10 and CIFAR100, the first convolutional layer uses a $3 \times 3$ kernel instead of $7 \times 7$.
For Cars196, CUB200-2011, and SOP, the network is initialized using weights pre-trained on ImageNet and all batch-normalization parameters are frozen.
We remove the head of the architecture and add two fully-connected layers directly following the global-average-pooling operation.
The first fully-connected layer maps from 2048 units to $n$ units, and the second fully-connected layer maps from $n$ units to $Y$ units.
The last fully-connected layer has no bias parameters and for \vmf, is initialized using the scheme detailed in Appendix \ref{sec:kappa_init}.
For CIFAR10 and CIFAR100, $n = 128$, and $Y = 10$ and $Y = 100$, respectively.
For Cars196, CUB200-2011, and SOP, $n = 512$, and $Y = 83$, $Y = 85$, and $Y = 9620$, respectively.

\subsection{Datasets}

Below we detail data preprocessing, data augmentation, and batch sampling for each of the datasets. We follow the preprocessing and augmentation procedures of \cite{Boudiaf2020} for Cars196, CUB200-2011, and SOP.
For batch sampling during training, we use an episodic scheme where we first sample $N$ classes at random and then sample $K$ instances from each of the $N$ classes.

\paragraph{MNIST \& FashionMNIST.} For MNIST and FashionMNIST, pixels are linearly scaled to be in $[0, 1]$. No data augmentation is used. The validation set is created by splitting off 15\% of the train data, stratified by class label.
For batch sampling, $N = 10$ and $K = 13$.

\paragraph{CIFAR10 \& CIFAR100.} During training, images are padded with 4 pixels on all sides via reflection padding, after which a $32 \times 32$ random crop is taken. Then, the image is randomly flipped horizontally and z-score normalized. Finally, each image is occluded with a randomly located $8 \times 8$ patch where occluded pixels are set to zero. During validation and testing, images are only z-score normalized.
The validation set is created by splitting off 15\% of the train data, stratified by class label.
For batch sampling, CIFAR10 has $N = 10$ and $K = 26$ and CIFAR100 has $N = 32$ and $K = 8$.

\paragraph{Cars196.} During training, images are resized to $256 \times 256$ and brightness, contrast, saturation, and hue are jittered randomly with factors in $[0.7, 1.3]$, $[0.7, 1.3]$, $[0.7, 1.3]$, and $[-0.1, 0.1]$, respectively.
A crop of random size in $[0.16, 1.0]$ of the original size and random location is taken and resized to $224 \times 224$.
Finally, the image is randomly flipped horizontally and z-score normalized.
During validation and testing, images are resized to $256 \times 256$, center-cropped to size $224 \times 224$, and z-score normalized.
The train set contains half of the classes and the test contains the other half.
The validation set is created by splitting off 15\% of the train classes.
For batch sampling, $N = 32$ and $K = 4$.

\paragraph{CUB200-2011.} During training, images are resized such that the smaller edge has size 256 while maintaining the aspect ratio.
Brightness, contrast, and saturation are jittered randomly, all with factors in $[0.75, 1.25]$. 
A crop of random size in $[0.16, 1.0]$ of the original size and random location is taken with the aspect ratio selected randomly in $[0.75, 1.33]$ and resized to $224 \times 224$.
Finally, the image is randomly flipped horizontally and z-score normalized.
During validation and testing, images are resized such that the smaller edge has size 256, center-cropped to size $224 \times 224$, and z-score normalized.
The train set contains half of the classes and the test contains the other half.
The validation set is created by splitting off 15\% of the train classes.
For batch sampling, $N = 32$ and $K = 4$.

\paragraph{SOP.} During training, images are resized to $256 \times 256$, a crop of random size in $[0.16, 1.0]$ of the original size and random location is taken with aspect ratio selected randomly in $[0.75, 1.33]$, and resized to $224 \times 224$.
Finally, the image is randomly flipped horizontally and z-score normalized.
During validation and testing, images are resized to $256 \times 256$, center-cropped to size $224 \times 224$, and z-score normalized.
The train set contains half of the classes and the test contains the other half.
The validation set is created by splitting off 15\% of the train classes.
For batch sampling, $N = 32$ and $K = 2$.

\subsection{Hyperparameters}
\label{sec:hyper}

Models were optimized with SGD and either standard momentum or Nesterov momentum.
Validation accuracy was monitored throughout training and if it did not improve for 15 epochs, the learning rate was cut in half.
If the validation accuracy did not improve for 35 epochs, model training was stopped early.
Model parameters were saved for the epoch resulting in the highest validation accuracy.
For fixed-set datasets, we found \arcface~was unstable during training caused by a degeneracy in the objective where embeddings were pushed to the opposite side of the hypersphere from the class weight vectors.
To fix the degeneracy, we introduced an additional hyperparameter for the number of epochs at the beginning of training where the margin, $m$, was set to zero. 
Once the network had trained for several epochs without the margin, we set the margin to a value greater than zero and trained \arcface~normally.
Here is a list of all hyperparameters with abbreviations:
\begin{multicols}{3}
\begin{itemize}
    \item Learning rate (LR)
    \item Temperature learning rate (TLR)
    \item Momentum (MOM)
    \item Nesterov momentum (NMOM)
    \item $\ell_2$ weight decay factor (WD)
    \item Margin for \arcface~($m$)
    \item Number of initial epochs with $m = 0$ ($m = 0$ Epochs)
    \item Curvature for \hyperbolic~($c$)
    \item $\lambda$ for \vmf~($\lambda$)
    \item Initial value of $\tau$ (Init $\tau$)
\end{itemize}
\end{multicols}

Hyperparameters were selected using a grid search.
The tables below specify the best hyperparameter values we found for each of the losses.
If a hyperparameter is not applicable for a loss, we mark the value with ``--''.

\clearpage

\begin{table}[h]
\begin{center}
\begin{tabular}{@{}ccccccccccc@{}}
\toprule
& LR & TLR & MOM & NMOM & WD & $m$ & $m = 0$ Epochs & $c$ & $\lambda$ & Init $\tau$ \\
\midrule
\standard & $0.1$ & -- & $0.9$ & False & $0.0$ & -- & -- & -- & -- & -- \\
\hyperbolic & $0.05$ & -- & $0.99$ & True & $0.0$ & -- & -- & $1 \times 10^{-5}$ & -- & -- \\
\spherical & $0.5$ & $0.001$ & $0.9$ & True & $0.0$ & -- & -- & -- & -- & $0.0$ \\
\arcface & $0.05$ & $0.001$ & $0.9$ & False & $0.0$ & $0.5$ & $20$ & -- & -- & $0.0$ \\
\vmf & $1.0$ & $0.001$ & $0.99$ & True & $0.0$ & -- & -- & -- & $0.4$ & $0.0$ \\
\bottomrule
\end{tabular}
\end{center}
\caption{Hyperparameter values for MNIST.}
\end{table}

\begin{table}[h]
\begin{center}
\begin{tabular}{@{}ccccccccccc@{}}
\toprule
& LR & TLR & MOM & NMOM & WD & $m$ & $m = 0$ Epochs & $c$ & $\lambda$ & Init $\tau$ \\
\midrule
\standard & $0.01$ & -- & $0.99$ & False & $0.0$ & -- & -- & -- & -- & -- \\
\hyperbolic & $0.1$ & -- & $0.9$ & True & $0.0$ & -- & -- & $1 \times 10^{-5}$ & -- & -- \\
\spherical & $0.5$ & $0.001$ & $0.9$ & True & $0.0$ & -- & -- & -- & -- & $0.0$ \\
\arcface & $0.01$ & $0.001$ & $0.99$ & True & $0.0$ & $0.5$ & $20$ & -- & -- & $0.0$ \\
\vmf & $0.05$ & $0.001$ & $0.99$ & False & $0.0$ & -- & -- & -- & $0.4$ & $0.0$ \\
\bottomrule
\end{tabular}
\end{center}
\caption{Hyperparameter values for FashionMNIST.}
\end{table}

\begin{table}[h]
\begin{center}
\begin{tabular}{@{}ccccccccccc@{}}
\toprule
& LR & TLR & MOM & NMOM & WD & $m$ & $m = 0$ Epochs & $c$ & $\lambda$ & Init $\tau$ \\
\midrule
\standard & $0.05$ & -- & $0.99$ & True & $1 \times 10^{-4}$ & -- & -- & -- & -- & -- \\
\hyperbolic & $0.01$ & -- & $0.99$ & True & $5 \times 10^{-4}$ & -- & -- & $1 \times 10^{-5}$ & -- & -- \\
\spherical & $0.5$ & $0.001$ & $0.8$ & False & $5 \times 10^{-4}$ & -- & -- & -- & -- & $0.0$ \\
\arcface & $0.1$ & $0.001$ & $0.9$ & True & $5 \times 10^{-4}$ & $0.4$ & $60$ & -- & -- & $0.0$ \\
\vmf & $0.5$ & $0.001$ & $0.9$ & False & $1 \times 10^{-4}$ & -- & -- & -- & $0.4$ & $0.0$ \\
\bottomrule
\end{tabular}
\end{center}
\caption{Hyperparameter values for CIFAR10.}
\end{table}

\begin{table}[h]
\begin{center}
\begin{tabular}{@{}ccccccccccc@{}}
\toprule
& LR & TLR & MOM & NMOM & WD & $m$ & $m = 0$ Epochs & $c$ & $\lambda$ & Init $\tau$ \\
\midrule
\standard & $0.1$ & -- & $0.9$ & True & $5 \times 10^{-4}$ & -- & -- & -- & -- & -- \\
\hyperbolic & $0.01$ & -- & $0.99$ & True & $5 \times 10^{-4}$ & -- & -- & $1 \times 10^{-5}$ & -- & -- \\
\spherical & $0.5$ & $0.0001$ & $0.8$ & False & $5 \times 10^{-4}$ & -- & -- & -- & -- & $0.0$ \\
\arcface & $0.01$ & $0.0001$ & $0.99$ & True & $5 \times 10^{-4}$ & $0.3$ & $100$ & -- & -- & $0.0$ \\
\vmf & $0.1$ & $0.01$ & $0.99$ & True & $1 \times 10^{-4}$ & -- & -- & -- & $0.4$ & $0.0$ \\
\bottomrule
\end{tabular}
\end{center}
\caption{Hyperparameter values for CIFAR100.}
\end{table}

\clearpage

\begin{table}[h]
\begin{center}
\begin{tabular}{@{}ccccccccccc@{}}
\toprule
& LR & TLR & MOM & NMOM & WD & $m$ & $m = 0$ Epochs & $c$ & $\lambda$ & Init $\tau$ \\
\midrule
\standard & $0.005$ & -- & $0.8$ & True & $5 \times 10^{-5}$ & -- & -- & -- & -- & -- \\
\hyperbolic & $0.01$ & -- & $0.9$ & True & $5 \times 10^{-4}$ & -- & -- & $1 \times 10^{-5}$ & -- & -- \\
\spherical & $0.005$ & $0.001$ & $0.9$ & True & $5 \times 10^{-4}$ & -- & -- & -- & -- & $3.466$ \\
\arcface & $0.0001$ & $0.0001$ & $0.9$ & False & $1 \times 10^{-4}$ & $0.3$ & $0$ & -- & -- & $3.466$ \\
\vmf & $0.0001$ & $0.01$ & $0.9$ & False & $1 \times 10^{-4}$  & -- & -- & -- & $0.7$ & $0.0$ \\
\bottomrule
\end{tabular}
\end{center}
\caption{Hyperparameter values for Cars196.}
\end{table}

\begin{table}[h]
\begin{center}
\begin{tabular}{@{}ccccccccccc@{}}
\toprule
& LR & TLR & MOM & NMOM & WD & $m$ & $m = 0$ Epochs & $c$ & $\lambda$ & Init $\tau$ \\
\midrule
\standard & $0.001$ & -- & $0.8$ & True & $5 \times 10^{-4}$ & -- & -- & -- & -- & -- \\
\hyperbolic & $0.005$ & -- & $0.9$ & False & $5 \times 10^{-4}$ & -- & -- & $1 \times 10^{-5}$ & -- & -- \\
\spherical & $0.001$ & $0.0001$ & $0.9$ & False & $5 \times 10^{-4}$ & -- & -- & -- & -- & $2.773$ \\
\arcface & $0.001$ & $0.0001$ & $0.8$ & True & $5 \times 10^{-4}$ & $0.3$ & $0$ & -- & -- & $2.773$ \\
\vmf & $0.001$ & $0.01$ & $0.9$ & False & $1 \times 10^{-4}$ & -- & -- & -- & $0.7$ & $2.773$ \\
\bottomrule
\end{tabular}
\end{center}
\caption{Hyperparameter values for CUB200-2011.}
\end{table}

\begin{table}[h]
\begin{center}
\begin{tabular}{@{}ccccccccccc@{}}
\toprule
& LR & TLR & MOM & NMOM & WD & $m$ & $m = 0$ Epochs & $c$ & $\lambda$ & Init $\tau$ \\
\midrule
\standard & $0.005$ & -- & $0.8$ & True & $5 \times 10^{-4}$ & -- & -- & -- & -- & -- \\
\hyperbolic & $0.0005$ & -- & $0.99$ & False & $1 \times 10^{-4}$ & -- & -- & $1 \times 10^{-5}$ & -- & -- \\
\spherical & $0.001$ & $0.0001$ & $0.99$ & True & $1 \times 10^{-4}$ & -- & -- & -- & -- & $0.0$ \\
\arcface & $0.005$ & $0.0001$ & $0.8$ & True & $5 \times 10^{-4}$ & $0.3$ & $0$ & -- & -- & $2.773$ \\
\vmf & $0.0001$ & $0.001$ & $0.8$ & True & $5 \times 10^{-4}$ & -- & -- & -- & $0.7$ & $2.773$ \\
\bottomrule
\end{tabular}
\end{center}
\caption{Hyperparameter values for SOP.}
\end{table}

\clearpage

\section{Temperature Details}
\label{sec:temp_details}

We parameterize the inverse-temperature, $\beta$, as $\beta = \exp(\tau)$ where $\tau \in \mathbb{R}$ is an unconstrained network parameter learned automatically via gradient descent.

For fixed-set classification tasks, the network is initialized randomly.
We find that fixing $\beta$ to a large value adversely affects the training dynamics, as small improvements in angular separation can lead to drastic reductions in the loss caused by the peakedness of the posterior (see the reduction in performance for a fixed value of $\beta$ in the left frame of Figure \ref{fig:fixed_temp}).
A challenge in learning $\tau$ directly is the network can cheat by maximizing it within the first several epochs.
Our parameterization prevents ``early cheating'' since smaller values of $\tau$ result in smaller gradients. 
For all fixed-set tasks, we initialize $\tau = 0$.
By using a smaller learning rate for $\tau$, the network is initially incentivized to focus on the angular discrimination between classes.

The backbone for the open-set tasks is a ResNet50 pretrained on ImageNet.
Because the network reuses previously learned features, we found initializing $\tau$ to a larger value sometimes results in improved performance. See the hyperparameter tables above for Cars196, CUB200-2011, and SOP for the initial values of $\tau$.

\section{Computing Infrastructure}

Each model was trained using a single NVIDIA Tesla v100 on Google Cloud Platform.
Code was implemented using PyTorch v1.6.0 and Python 3.8.2 on Ubuntu 18.04.

\section{Runtime of the von Mises--Fisher Loss}
\label{sec:runtime}

\vmf~does require rejection sampling, but in practice the sampling has marginal impact on runtime.
For Cars196, training is 75 seconds per epoch for both \vmf~and \spherical, and 32 seconds vs. 28 seconds for computing test-set embeddings.
For CIFAR100, \vmf~is slightly slower to train than \spherical, 35 seconds per epoch vs. 28 seconds per epoch, but computing
test-set predictions is 3 seconds for both.
\vmf~is no more sensitive to hyperparameters, but it does require slightly
more epochs to converge: for Cars196 and CIFAR100, \vmf~trained for 1.8x and 1.1x as many epochs compared to \spherical, respectively.

\section{Cars196 Test Images}

Below we show paired grids of Cars196 test images for each of the losses where the left grid contains images corresponding to the smallest $\norm{\bz}$ and the right grid corresponding to the largest $\norm{\bz}$. The provided figures are analogous to Figure \ref{fig:vmf_cars}. Note that for \vmf, $\norm{\bzt} = \kappa_{\bz}$.

\begin{figure}[h]
\centering
\includegraphics[width=0.9\textwidth]{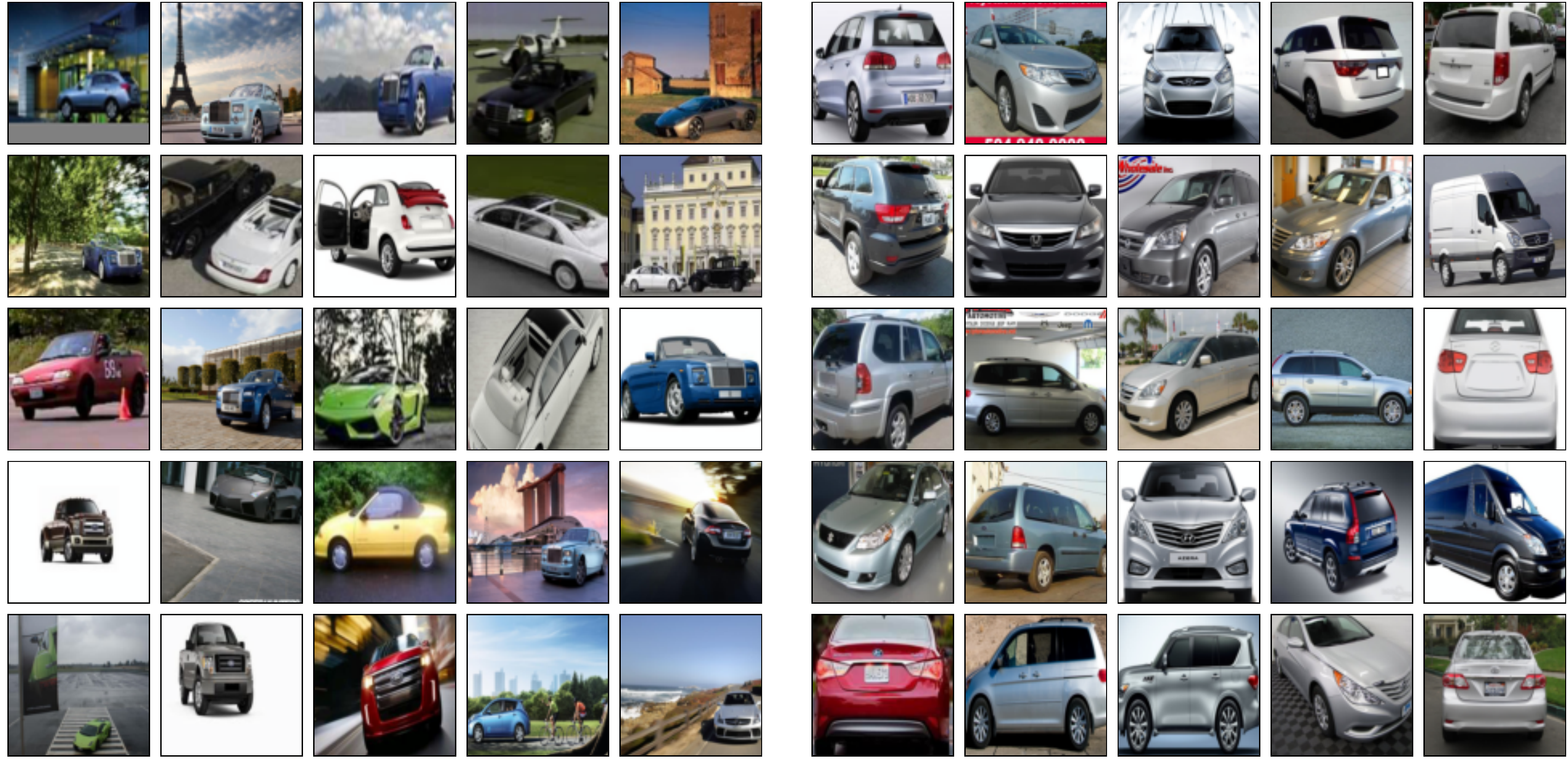}
\caption{\standard~embeddings with the (left) smallest $\norm{\bz}$ and (right) largest $\norm{\bz}$.}
\end{figure}

\begin{figure}[h]
\centering
\includegraphics[width=0.9\textwidth]{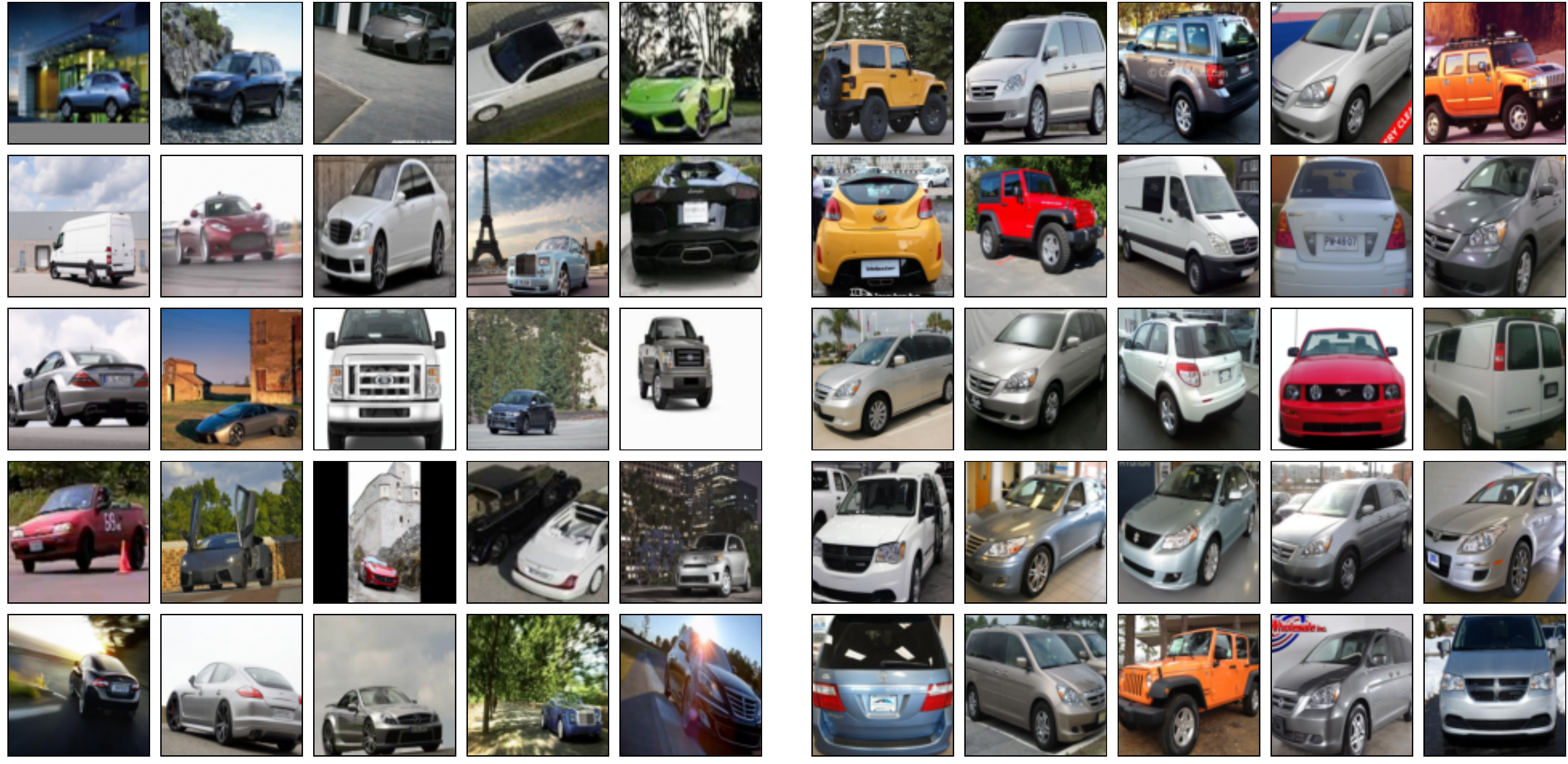}
\caption{\hyperbolic~embeddings with the (left) smallest $\norm{\bz}$ and (right) largest $\norm{\bz}$.}
\end{figure}

\begin{figure}[h]
\centering
\includegraphics[width=0.9\textwidth]{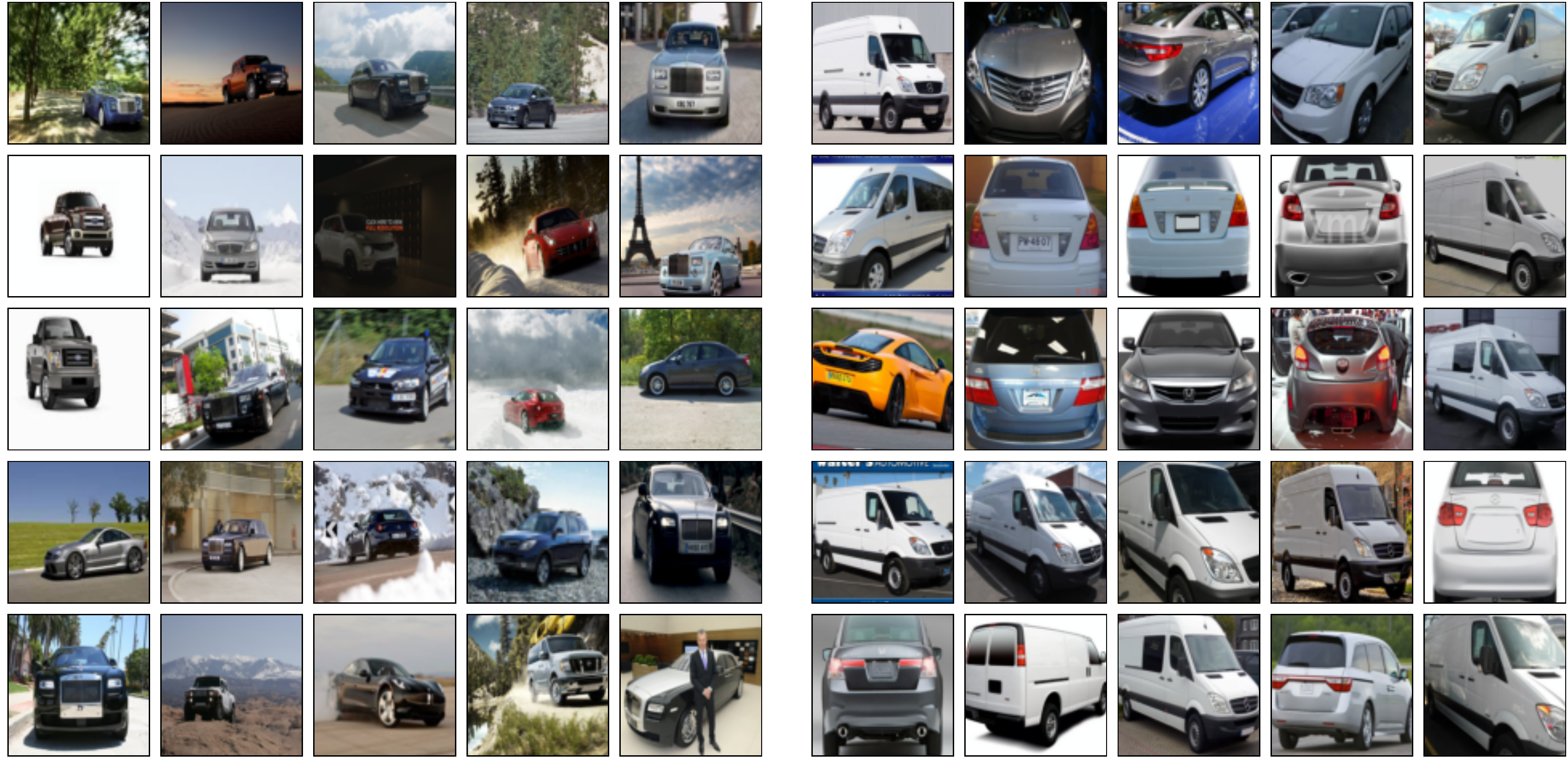}
\caption{\spherical~embeddings with the (left) smallest $\norm{\bz}$ and (right) largest $\norm{\bz}$.}
\end{figure}

\begin{figure}[h]
\centering
\includegraphics[width=0.9\textwidth]{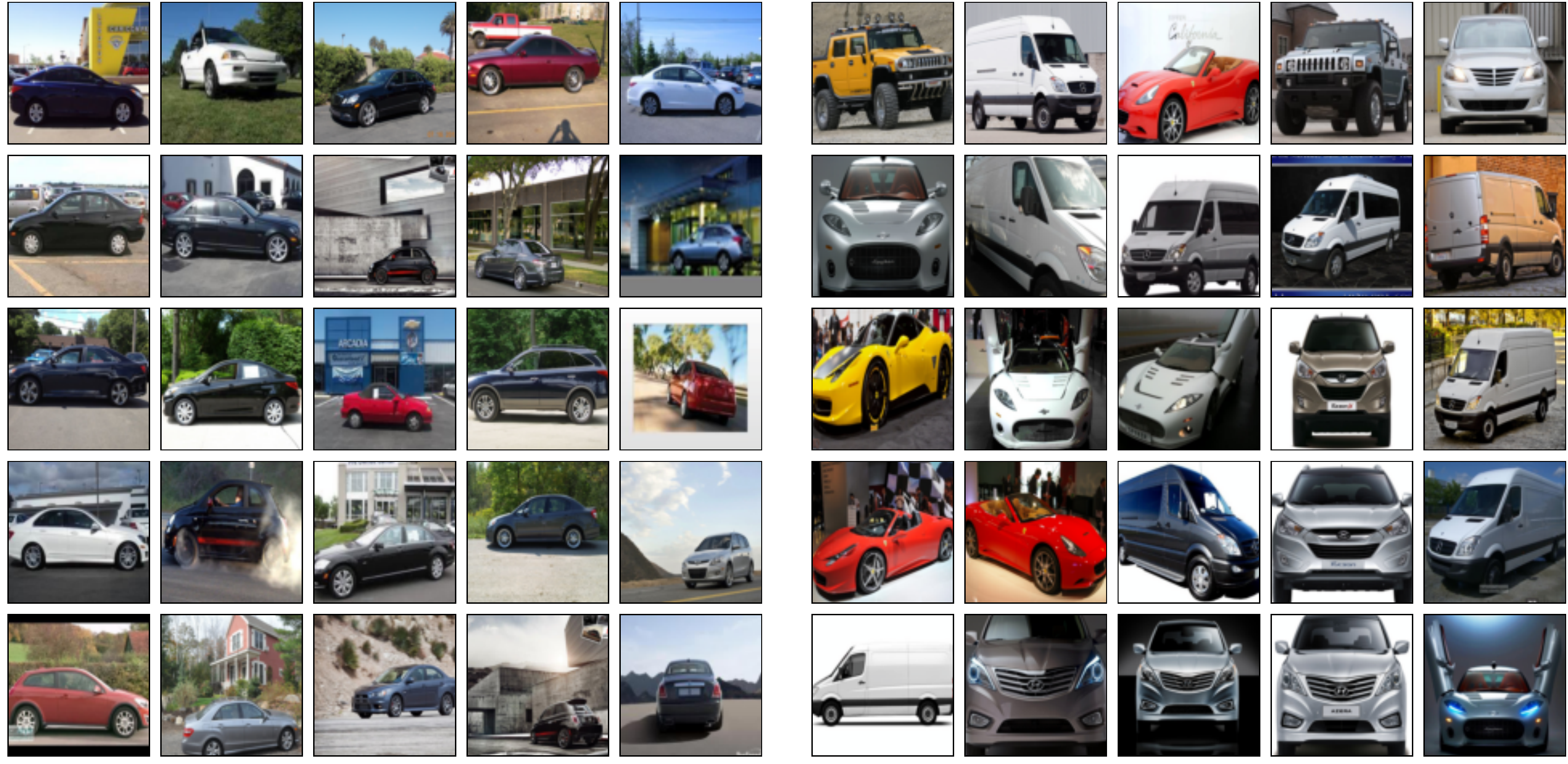}
\caption{\arcface~embeddings with the (left) smallest $\norm{\bz}$ and (right) largest $\norm{\bz}$.}
\end{figure}

\begin{figure}[h]
\centering
\includegraphics[width=0.9\textwidth]{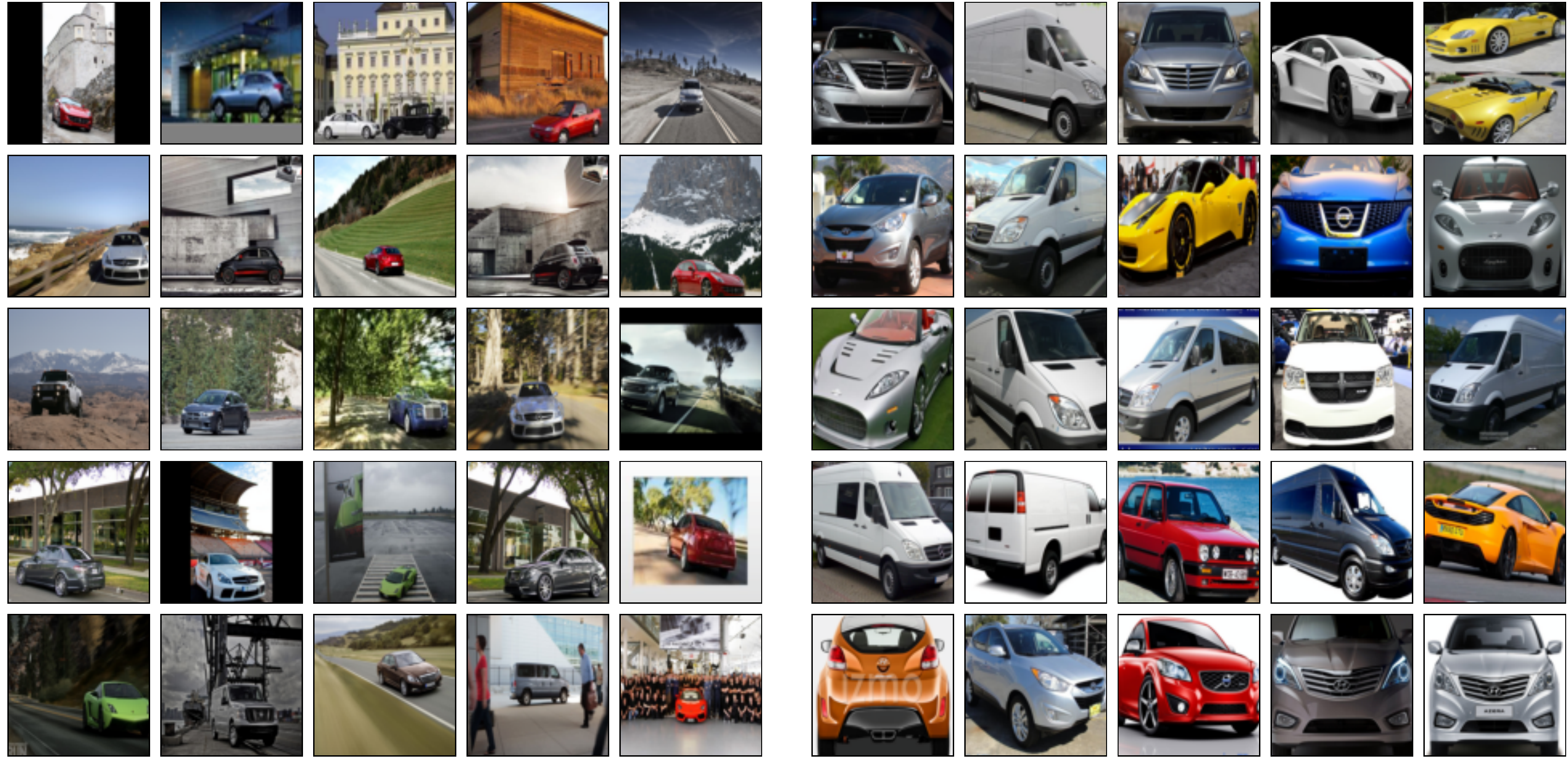}
\caption{\vmf~embeddings with the (left) smallest $\kappa_{\bz}$ and (right) largest $\kappa_{\bz}$.}
\end{figure}

}

\end{document}